\documentclass[10pt,twocolumn,letterpaper]{article}

\usepackage{cvpr}              

\usepackage{marvosym}
\usepackage{amssymb}
\usepackage{graphicx}
\usepackage{amsmath}
\usepackage{booktabs}
\usepackage{rotating}
\usepackage{multirow}
\usepackage{enumitem}
\usepackage{algorithmicx,algorithm}
\usepackage{enumitem}
\usepackage{pifont}
\usepackage{makeidx}
\usepackage{epstopdf}
\usepackage[noend]{algpseudocode}
\usepackage{xcolor}
\usepackage{colortbl}
\usepackage{color}

\usepackage[pagebackref,breaklinks,colorlinks]{hyperref}
\usepackage[capitalize]{cleveref}
\crefname{section}{Sec.}{Secs.}
\Crefname{section}{Section}{Sections}
\Crefname{table}{Table}{Tables}
\crefname{table}{Tab.}{Tabs.}

\colorlet{dark-blue}{blue!70!black}
\hypersetup{
    colorlinks=true,%
    citecolor=dark-blue,%
    filecolor=dark-blue,%
    linkcolor=red,%
    urlcolor=magenta
}

\def\cvprPaperID{7100} 
\def\confName{CVPR}
\def\confYear{2024}

\makeatletter
\def\thanks#1{\protected@xdef\@thanks{\@thanks
        \protect\footnotetext{#1}}}
\makeatother

\begin{document}

\title{Stable Neighbor Denoising for Source-free Domain Adaptive Segmentation}

\thanks{ 
This work was supported in part by the National Natural Science Foundation of China No. 62271377, the Key Research and Development Program of Shannxi Program No. 2021ZDLGY0106 and No. 2022ZDLGY0112, the National Key R\&D Program of China under Grant 2021ZD0110400 and 2021ZD0110404 the Key Scientific Technological Innovation Research Project by Ministry of Education, the Joint Funds of the National Natural Science Foundation of China (U22B2054).    \\
\quad \quad \textsuperscript{\Letter} Co-corresponding author.}

\author{Dong Zhao$ ^{1}$, Shuang Wang$^{1}$ \textsuperscript{\Letter}, Qi Zang$ ^{1}$, Licheng Jiao$ ^{1}$, Nicu Sebe$ ^{2}$, Zhun Zhong$ ^{3,4}$ \textsuperscript{\Letter}\\
$ ^{1}$ School of Artificial Intelligence,
Xidian University, Shaanxi, China \\
$ ^{2}$ Department of Information Engineering and Computer Science, University of Trento, Italy \\
$ ^{3}$ School of Computer Science and Information Engineering, Hefei University of Technology, China \\
$ ^{4}$ School of Computer Science, University of Nottingham, NG8 1BB Nottingham, UK \\
}
\maketitle

\begin{abstract}
\vspace{-0.2cm}

We study source-free unsupervised domain adaptation (SFUDA) for semantic segmentation, which aims to adapt a source-trained model to the target domain without accessing the source data.
Many works have been proposed to address this challenging problem, among which uncertainty-based self-training is a predominant approach.
However, without comprehensive denoising mechanisms, they still largely fall into biased estimates when dealing with different domains and confirmation bias.
In this paper, we observe that pseudo-label noise is mainly contained in unstable samples in which the predictions of most pixels undergo significant variations during self-training. 
Inspired by this, we propose a novel mechanism to denoise unstable samples with stable ones.
Specifically, we introduce the Stable Neighbor Denoising (SND) approach, which effectively discovers highly correlated stable and unstable samples by nearest neighbor retrieval and guides the reliable optimization of unstable samples by bi-level learning.
Moreover, we compensate for the stable set by object-level object paste, which can further eliminate the bias caused by less learned classes. 
Our SND enjoys two advantages. 
First, SND does not require a specific segmentor structure, endowing its universality. 
Second, SND simultaneously addresses the issues of class, domain, and confirmation biases during adaptation, ensuring its effectiveness. 
Extensive experiments show that SND consistently outperforms state-of-the-art methods in various SFUDA semantic segmentation settings. 
In addition, SND can be easily integrated with other approaches, obtaining further improvements.
The source code is available at \href{https://github.com/DZhaoXd/SND}{https://github.com/DZhaoXd/SND}.
\end{abstract}

\vspace{-0.2cm}
\section{Introduction}
\label{sec:intro}
\vspace{-0.2cm}

Unsupervised domain adaptation (UDA) \cite{pan2009survey} transfers knowledge of the labeled-rich source domains to the unlabeled target domains, providing an effective solution for semantic segmentation towards lower annotation burden \cite{hoffman2018cycada} and strong generalization in an open environment\cite{xie2023sepico, 10187687}.
However, the access requirements of source domain data make it impossible to handle scenarios involving privacy, property rights protection, and confidentiality\cite{fleuret2021uncertainty}.
Thus, this work focuses on a more practical task in semantic segmentation, source-free unsupervised domain adaptation  (SFUDA)\cite{yang2021generalized, liu2021source}, which aims to adapt a source-trained model to a target domain without accessing the source data.

\begin{figure}[tbp]
    \begin{center}
    \centering \includegraphics[width=0.480\textwidth, height=0.32\textwidth]{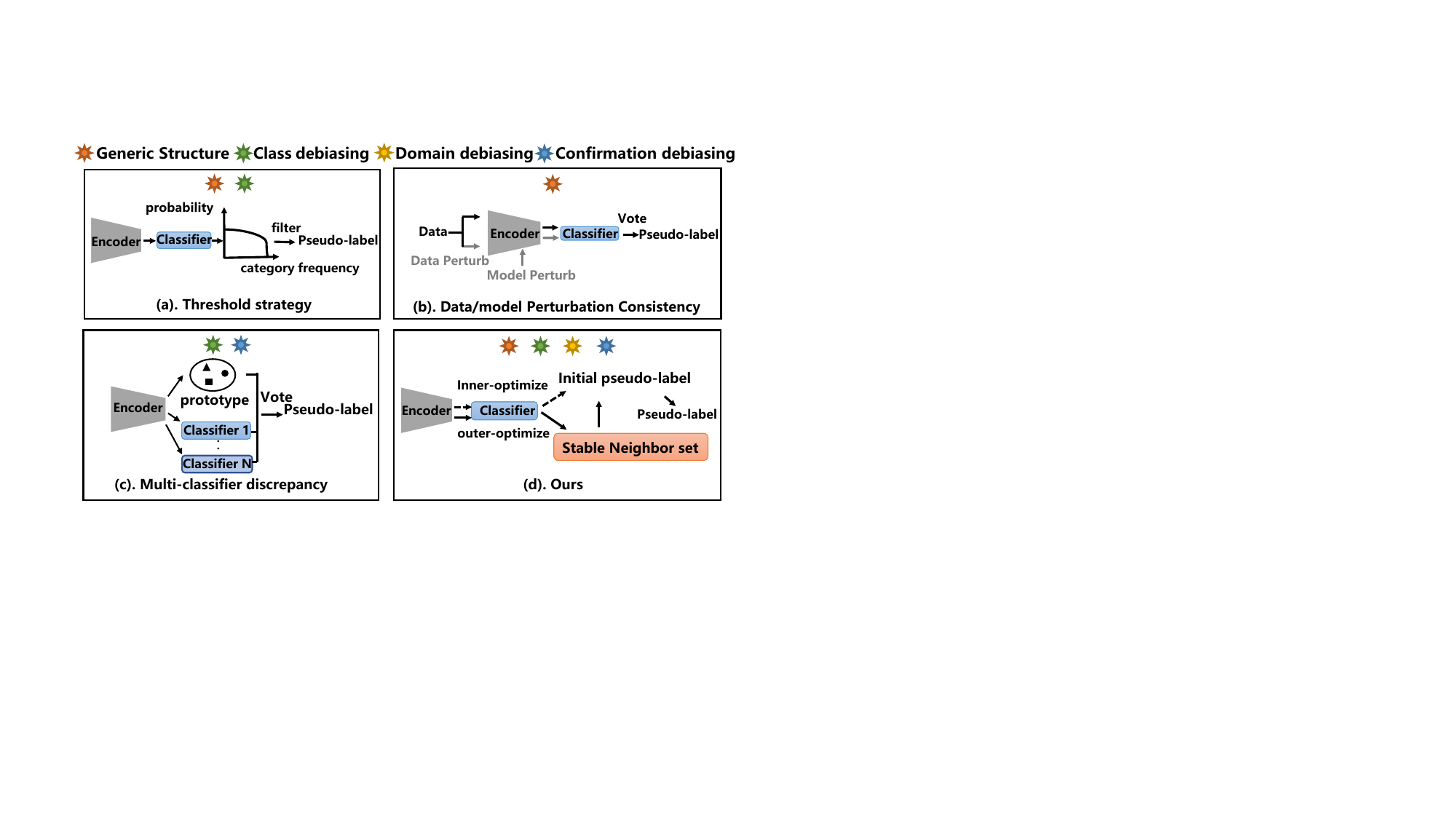}
    \end{center}
    \setlength{\abovecaptionskip}{-0.1 cm}
    \caption{Comparison of advantages of different uncertain estimation strategies in self-training method for SFUDA.} 
    \label{advantage}
\vspace{-0.45cm}
\end{figure}

Currently, self-training is the mainstream technology to address the SFUDA problem in semantic segmentation, which strives to adapt the model with high-quality pseudo-labels produced by uncertain estimation functions, \emph{e.g.},  probability thresholding \cite{araslanov2021self, li2022class, zhao2023towards, hoyer2023mic, hoyer2022hrda, hoyer2022daformer, ijcv_org}, perturbation consistency \cite{fleuret2021uncertainty, Kundu_2021_ICCV, yang2023exploring}, or discrepancy classifier voting \cite{zheng2021rectifying, zhang2021prototypical, chen2022deliberated, shen2023diga, Kundu_2021_ICCV, 10181233}.
Generally, a strong cross-domain self-training method should comprehensively consider the following aspects.
\emph{\ding{182} Generic structure}: It does not require the adoption of a specific structural design (e.g., a specific segmentation head) on the source side; otherwise, it would not be conducive to target deployment~\cite{pan2023cloud}.
\emph{\ding{183} Class debiasing}:  the less-learned (hard) categories in the source model should be paid more attention, as the class imbalance problem often exists in segmentation tasks~\cite{araslanov2021self, hoyer2022daformer, tranheden2021dacs, zou2018unsupervised, ma2023delving}. 
\emph{\ding{184} Domain debiasing}: the self-training method needs to maintain the adaptability of distinct domains since the cross-domain segmentation tasks often face multiple or compound domains, \textit{e.g.}, different weather conditions.
\emph{\ding{185} Confirmation debiasing}: since the running confidence of the model will increase with self-training~\cite{chen2022debiased, Zhao_2023_ICCV}, a strong self-training method should be able to dynamically adjust the uncertain estimation function for addressing the confirmation bias.
However, existing self-training methods for SFUDA, lacking a comprehensive denoising mechanism, often fall short in one or more aspects (see Fig.~\ref{advantage}(a-c)), leading to inefficiency and under-adaptation.

\begin{figure}[tbp]
    \begin{center}
    \centering 
    \includegraphics[width=0.40\textwidth, height=0.32\textwidth]{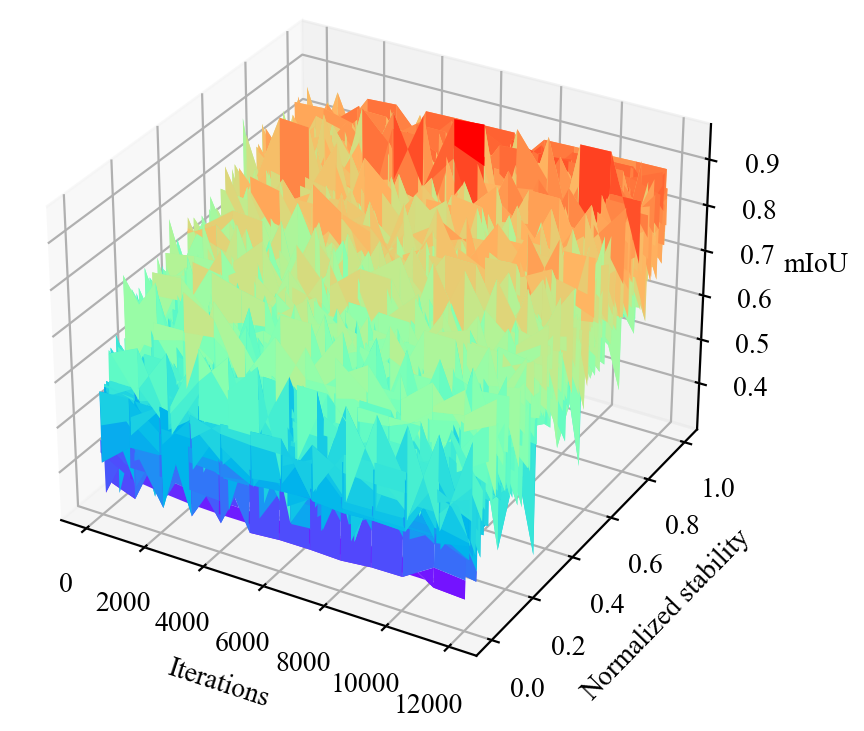}
    \end{center}
    \setlength{\abovecaptionskip}{-0.1 cm}
    \caption{The plot of the mIoU scores versus stability for each target sample throughout training. Stability is calculated by the difference between the initial and the current segmentation. It shows a positive correlation between mIoU (pseudo-label quality) and stability. This observation holds during the whole training process. Experiments are from the GTA5 $\rightarrow$ Cityscapes SFUDA task.}
    \label{stability}
\vspace{-0.45cm}
\end{figure}

Our motivation comes from experimental observations of implementing the vanilla self-training~\cite{araslanov2021self} on SFUDA. As shown in Fig.~\ref{stability}, for each target training sample at different iterations, we record 1) its mIoU scores with the ground truth (mIoU-axis); and 2) the change degrees between the segmentation results of the current step and the ones obtained by the initial step (stability-axis). 
Throughout the training process, we observe that samples with high stability consistently maintain high mIoU scores, while those with low stability consistently show low mIoU scores.
This indicates that noise in samples with low stability is the main contributor to error accumulation during self-training. 
Moreover, this issue does not significantly improve as training progresses. 
This motivates us to identify stable samples (\emph{i.e.}, whose pseudo-label change degree are low) and unstable samples to address the above issue.

To this end, we propose a novel denoising approach, called \emph{Stable Neighbor Denoising} (SND), which can effectively discover highly correlated stable and unstable samples by nearest neighbor retrieval and denoise unstable samples with the guidance of the stable ones by bi-level optimization. 
Specifically, we resort to bi-level optimization \cite{wu2021learning, Pham_2021_CVPR} to establish the connection between the stable and unstable sets and guide the evolution direction of unstable samples.
The principle here is that if the model is trained on unstable samples with correct pseudo-labels, the loss on stable samples will be also minimized.
Importantly, by the analysis of the gradient of bi-level optimization (detailed in Section~\ref{relations}), we find that matching unbiased and highly correlated stable samples for unstable samples is the key to achieving this principle.
This is because the loss gradients of biased/low-correlation samples often have non-intersecting/different optimization directions, leading to bias in bi-level optimization for denoising. 
To tackle this problem, we propose retrieving stable neighbors by image style and spatial layout factors to eliminate the domain bias in one-step bi-level denoising optimization. In addition,we propose refining the stable samples into two sets: a whole-stable set and a category-stable set. Then, we compensate for stable neighbors using diverse object-level pasting to eliminate category bias.
In this way, bi-level optimization can efficiently denoise unstable samples with stable ones under the principle.

Through the above techniques, the proposed SND shows superiority over previous works as follows. 
\emph{\ding{182} Generic structure}: 
SND does not require a specific network structure on both the source and target sides, endowing its universality.
\emph{\ding{183} Class debiasing}: SND constructs stable neighbors with balanced categories and uses them to guide the adaptation of less-learned categories.
\emph{\ding{184} Domain debiasing}:  SND denoises distinct domains by retrieving neighbors with similar domain factors and can handle complex compound domain scenarios (See Table.~\ref{abl_ue}).
\emph{\ding{185} Confirmation debiasing}: SND designs a stability-oriented denoising mechanism that filters out noisy pseudo-labels without the need for probability or prototype distance, thereby mitigating confirmation bias (See Fig.~\ref{train_cur}).
Extensive experiments show that SND consistently outperforms state-of-the-art methods in various SFUDA semantic segmentation settings. 

\vspace{-0.1cm}
\section{Related Work}
\vspace{-0.1cm}

\noindent \textbf{Unsupervised Domain Adaptation}. Domain adaptive semantic segmentation has achieved significant adaptability improvements in multiple adaptation scenarios such as synthesis-to-reality\cite{hoyer2023mic, hoyer2022hrda, hoyer2022daformer, cheng2023adpl, shen2023diga}, cross-weather\cite{sakaridis2020map, sakaridis2019guided, lee2022fifo}, multi-source\cite{wu2021dannet, he2021multi}, etc.
Overall, the current work mainly realizes cross-domain transfer from the following aspects. 1) Use the source domain data to learn a generalized representation\cite{chen2024liebn}. This line of work employs data augmentation (e.g. copy-paste\cite{tranheden2021dacs}, random masking\cite{hoyer2023mic}) and domain randomization (e.g. expanding image styles \cite{huang2021fsdr, kundu2021generalize, shen2023diga, choi2021robustnet, yue2019domain, zhao2022style}) to expand the representation space of source domain models. 2) Align the source and target domains. The following works adopt a variety of domain alignment strategies (e.g. adversarial training or statistical matching) in a variety of alignment spaces (e.g. pixels space\cite{hoffman2018cycada, chen2019crdoco}, Fourier space\cite{Yang_2020_CVPR}, features space\cite{FADA_ECCV_2020, pu2022meta, 9961139}, output space\cite{vu2019advent, Luo2019TakingAC, Tsai_2019_ICCV}, etc.), reducing the statistical difference of both domains. 3) Target characteristic mining. Most of such works use augmentation consistency \cite{choi2019self, zhang2021prototypical, Araslanov_2021_CVPR, Ma_2023_CVPR} and pseudo-labels \cite{MaxSquare_2019_ICCV, yue2019domain} to further improve the model's adaptive ability, \emph{e.g.} tail category and local distribution.

\noindent \textbf{Source-Free Unsupervised Domain Adaptation} (SFUDA).
In classification tasks, prior SFUDA works propose implicit distribution alignment\cite{liang2021source, ding2022source}, instance contrast\cite{zhang2022divide}, and model perturbation\cite{jing2022variational} to align domains without source data.
In segmentation tasks, the above idea is hard to adopt due to complex semantic feature spaces. 
Most SFUDA works for segmentation adopt self-training, which filters and retrains the well-adapted pixels or regions through probability thresholds\cite{zhao2023towards}, data \cite{ huang2021model} or model discrepancy\cite{fleuret2021uncertainty, Kundu_2021_ICCV}, but they often fall into various biases.

\noindent \textbf{Noisy Label Learning}.
SFUDA can be viewed as a noisy label learning problem\cite{yi2023source, yang2022divide, zhao2023semantic}. 
Currently, noisy label learning is mostly discussed on classification tasks. 
The mainstream technologies include robust loss design\cite{zhang2018generalized, wang2019symmetric, ma2020normalized}, self-label correction\cite{cheng2020learning, li2022expansion}, prototype denoising\cite{han2019deep}, meta-learning based denoising\cite{wu2021learning, Pham_2021_CVPR}, etc. 
Most of those methods are designed for instance-level and are not suitable for pixel-level segmentation tasks.
In particular, the prototype \cite{zhang2021prototypical, li2022weakly} and meta-learning \cite{MetaCorrection} based denoising methods have been applied to UDA segmentation tasks, but their dependence on source data limits on SFUDA.

\vspace{-0.1cm}
\section{Method}
\vspace{-0.1cm}
\label{sec:method}

\noindent \textbf{Preliminary.} In the setting of source-free unsupervised domain adaptation (SFUDA), we are given a segmentation model $ \mathcal H$ with parameter $ \Theta $ pre-trained on the labeled source dataset $ \ {\mathcal D}_{s}=\{(x^{i}_{s}, y^{i}_{s})\}_{i}^{N_{s}} $, and the unlabeled target dataset $ \ {\mathcal D}_{t}=\{x^{i}_{t}\}_{i}^{N_{t}} $.
The goal is to adapt the network $ \mathcal H	$ and achieve low risk on the target dataset without accessing the source data $ \ {\mathcal D}_{s}$.
To achieve this goal, most works mainly conduct self-training to optimize the $ \mathcal H	$ as follows,
\begin{equation} \label{base_certain}
\Theta^{\star	} = \mathop{\arg\min}_{\Theta} \sum_{i}^{N_t}  \sum_{l}^{H \times W} 
 \omega^{(i,l)}\mathcal L[\mathcal H(x_{t}^{(i,l)}  |  \Theta ), \hat y_{t}^{(i,l)}],
\end{equation}
where $\mathcal L$ is the cross-entropy loss and $\hat y_{t}$ is the pseudo-label.  $\omega$ is a weighting factor calculated by uncertainty, in which the lower the uncertainty, the closer the value is to 0.
This paper proposes stable neighbor denoising techniques to estimate $\omega$ in an unbiased way.

\subsection{Division of Stable \& Unstable Sets}
\vspace{-0.1cm}

Following the observation in Fig \ref{stability}, we aim to divide target domain samples into stable and unstable sets by the change degree (also called evolution stability) of their segmentation results during the vanilla self-training \cite{araslanov2021self}. 
Formally, for each target sample $x_t^i$, we define its evolution stability (${\rm ES}$) at $\tau$-th iteration as follow,
\vspace{-0.1cm}
\begin{equation} \label{es}
{\rm ES}^{i, \tau} =  \sum_{l}^{H \times W} 
\mathcal {\rm SIM}  [\mathcal H(x_{t}^{i, l}  |  \Theta^{0}), \mathcal H(x_{t}^{i, l}  |  \Theta^{\tau})],
\vspace{-0.1cm}
\end{equation} 
where ${\rm SIM}[\cdot]$ is the cosine similarity. $\Theta^{\tau}$ represent the parameters of the model at $\tau$-th iteration during self-training.
The larger the ${\rm ES}$ value of the sample, the more stable it is.
As ${\rm ES}$ may fluctuate in different degrees of domain shift tasks, it is hard to set a suitable threshold to determine whether it is stable.
Instead, we take the top-$k\%$ ranked highly stable samples to divide the $ \ {\mathcal D}_{t}$ into the stable set ${\mathcal D}_{se}$ and the unstable set ${\mathcal D}_{ue}$ as follows, 
\begin{equation} \label{d_se}
 \ {\mathcal D}_{se}=\{(x_t^{i}, \bar y^{i}) | x_t^{i} \in  \ {\mathcal D}_{t}, { \rm ES}^{i,\tau} \in  \top_k \}, 
\end{equation}
\begin{equation} \label{d_ue}
\ {\mathcal D}_{ue}=\{x_t^{i}|  x_t^{i} \in  \ {\mathcal D}_{t} ,{ \rm ES}^{i,\tau} 	\notin  \top_k \} .
\end{equation}
Next, the main challenge is how to utilize $ \ {\mathcal D}_{se}$ to assist in predicting unbiased $\omega$ of $ \ {\mathcal D}_{ue}$.

\begin{figure*}[tbp]
    \begin{center}
    \centering 
    \includegraphics[width=0.925\textwidth, height=0.315\textwidth]{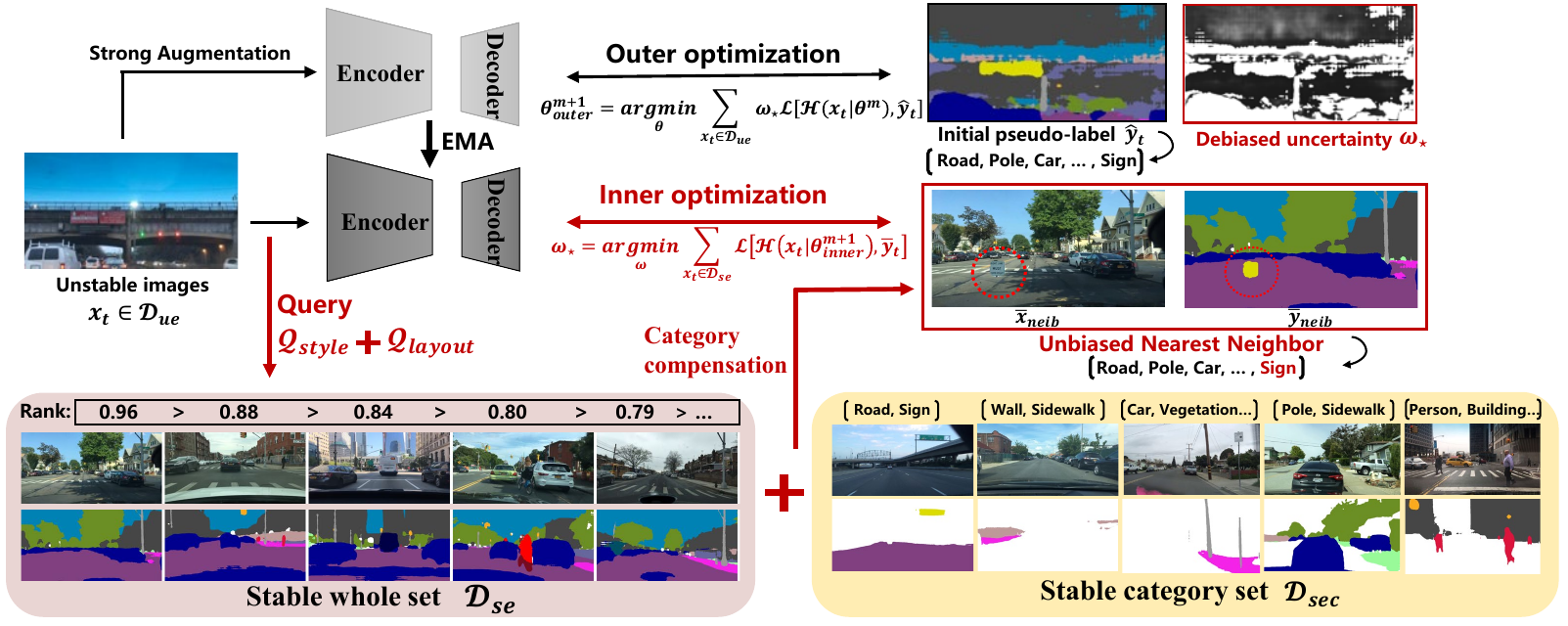}
    \end{center}
    \setlength{\abovecaptionskip}{-0.1 cm}
    \caption{The pipeline of the proposed \emph{Stable Neighbor Denoising} (SND). 
    It is formed by the student-teacher model\cite{sohn2020fixmatch}. 
    In each optimization, SND performs the inner and outer optimizations sequentially. 
    In the inner loop (\textcolor[RGB]{192,0,0}{red line}), SND utilizes the style $\mathcal Q_{style}$ and the layout $\mathcal Q_{layout}$ factors to retrieve stable neighbors for unstable samples and then performs category compensation to reduce category bias. 
    Thereafter, SND executes Eq. \ref{meta} using the teacher model to obtain the unbiased uncertainty map $\omega_{\star}$ and initial pseudo-label $\hat y_{t}$.
    In the outer loop (black line), SND performs Eq. \ref{outer_1} to optimize the student model. 
    EMA denotes the Exponential Moving Average. } 
    \label{pipeline}
\vspace{-0.3cm}
\end{figure*}
\subsection{Build Relations between Stable \& Unstable Sets} \label{relations}
An intuitive way is to exploit the cross-image feature similarity \cite{10005033} of samples in the two sets to build relations.
However, we find that the un-adapted source-trained model fails to capture this cross-image feature similarity because the representation ability of deep networks is always insufficient under the domain shift (see Table \ref{abl_ue}).
Instead, we propose to build relations between the two sets from an optimization perspective and utilize bi-level learning \cite{wu2021learning, Pham_2021_CVPR} to achieve this.
In a nutshell, we formulate each optimization step as the inner- and outer loops, in which the former learns uncertainty parameter $\omega$ for the latter.

Specifically, for $m$-th optimization step, in the inner loop, the optimizable $\omega$ is first initialized as $\omega_0$ and the model with parameters $ \Theta^{m}$ is optimized on unstable samples by Eq.\ref{base_certain}, 
\vspace{-0.2cm}
\begin{equation}
\Theta^{m+1}_{inner} = \mathop{\arg\min}_{\Theta} \sum_{\tiny { x_{t} \in  \ {\mathcal D}_{ue}} } \sum_{l}^{H \times W} 
\mathcal \omega^{i}_{0} \mathcal L[\mathcal H(x_{t}^{(i,l)}  |  \Theta^{m} ), \hat y_{t}^{(i,l)}]. \label{inner_0}
\end{equation}
$\omega^{i}_{0}$ is simply set to be an all-one matrix. 
For optimization, Eq.~\ref{inner_0} can be approximated by one- or multi-step gradient calculation, \emph{i.e.},  $\Theta^{m+1}_{inner} = \Theta^{m} - \alpha \sum_{\tiny { x_{t} \in  \ {\mathcal D}_{ue}} } \omega^{i}_{0} \frac{\partial \mathcal L(x_t^i | \Theta^{m} ) }{\partial \Theta} |_{\Theta^m}, $
where $\alpha$ is the inner-loop learning rate.
It can be seen that $\omega$ is differentiable relative to $ \Theta^{m+1}_{inner}$.
Subsequently, we optimize the $\omega$ by enforcing the optimized model with parameters $ \Theta^{m+1}_{inner}$ also achieves low risk on the stable samples in $ \ {\mathcal D}_{se}$,  
\vspace{-0.2cm}
\begin{equation}
\omega_{\star	} = \mathop{\arg\min}_{\omega} \sum_{\tiny { (x_{t},\bar y_{t}) \in  \ {\mathcal D}_{se}} }  \sum_{l}^{H \times W} 
\mathcal L[\mathcal H(x_{t}^{(i, l)}  |  \Theta^{m+1}_{inner} ), \bar y_{t}^{(i,l)}]. \label{inner_1}
\end{equation}

This object is driven by the following \emph{principle}: $\omega$ is optimized to increase/decrease the weight of these regions so that the model optimized on the unstable set samples has a small/large loss on the stable set.
In this way, we establish an implicit relation between stable and unstable sets, because optimizing noisy regions of unstable samples has misleading optimization goals.
Finally, in the outer loop, the model can be optimized by the learned $\omega_{\star}$,
\vspace{-0.2cm}
\begin{equation}
\Theta^{m+1}_{outer} = \mathop{\arg\min}_{\Theta} \sum_{\tiny { x_{t} \in  \ {\mathcal D}_{ue}} } \sum_{l}^{H \times W} 
\mathcal \omega_{\star} \mathcal L[\mathcal H(x_{t}^{(i,l)}  |  \Theta^{m} ), \hat y_{t}^{(i,l)}]. \label{outer_1}
\end{equation}
\noindent \textbf{Problem Discussion}. One important question in the above learning strategy is, does this \emph{principle} work by taking arbitrary samples from the stable and unstable set to optimize?
We present further analysis of the optimization of $ \omega$ to answer this question.
For optimization, Eq.~\ref{inner_1} is performed by gradient descent as follows, 
\begin{small}
\begin{align} \label{meta}
\omega_{\star}& = \omega_{0} - \beta \sum_{\tiny { x_{t} \in  \ {\mathcal D}_{se}} } \frac{\partial \mathcal L(x_t^i | \Theta^{m+1}_{inner} ) }{\partial \omega} |_{\omega_0}  \nonumber \\
& = \omega_{0} - \beta \sum_{\tiny { x_{t} \in  \ {\mathcal D}_{se}} } \frac{\partial \mathcal L(x_t^i |\Theta^{m+1}_{inner} ) }{\partial \Theta} |_{\Theta^{m+1}_{inner}} \cdot  \frac{\partial \Theta^{m+1}_{inner}}{\partial \omega} |_{\omega_0} \nonumber \\
 & =  \omega_{0} \ + \nonumber \\
 & \beta  \underbrace{ \sum_{\tiny { x_{t} \in  \ {\mathcal D}_{se}} }  \frac{\partial \mathcal L(x_t^i | \Theta^{m+1}_{inner} ) }  {\partial \Theta} |_{\Theta^{m+1}_{inner}}}_{\text{Loss gradient of stable set}}  \cdot \alpha \underbrace{ \sum_{\tiny { x_{t} \in  \ {\mathcal D}_{ue}} } \frac{\partial \mathcal L(x_t^i | \Theta^{m} ) }{\partial \Theta} |_{\Theta^m}}_{\text{Loss gradient of unstable set}} . \nonumber \\  
\end{align}
\end{small}


\begin{figure}[tbp]
    \begin{center}
    \centering 
    \includegraphics[width=0.48\textwidth, height=0.16\textwidth]{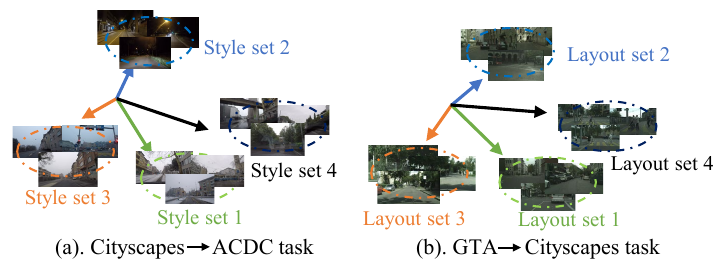}
    \end{center}
    \setlength{\abovecaptionskip}{-0.1 cm}
    \caption{Loss gradient directions in bi-level optimization.
    It is shown by computing the average gradient direction of different sets of classifier weights (ASPP) using the true labels.
    The black lines are inner-loop and the other colored lines are outer-loop.} 
    \label{dir}
\vspace{-0.45cm}
\end{figure}
Thus, when the continuous loss gradient on the stable samples in ${\mathcal D}_{se}$ has a similar direction to that on the unstable samples in ${\mathcal D}_{ue}$, the $\omega$ value of the corresponding area of the unstable samples will increase, and vice versa.
This indicates that optimizing noisy pseudo-labels would deviate from the optimization goal of stable samples and thus explains why noise in pseudo-labels can be estimated.
Moreover, Eq.~\ref{meta} also gives us the answer that arbitrarily selecting samples from the two sets will induce bias in the estimation on $\omega$, especially for samples with low correlation. This is because heterogeneous data always enjoy different optimal weight spaces~\cite{izmailov2018averaging}, which will mislead the estimation of $\omega$. We provide below a detailed analysis.

 \noindent \textbf{Reason \& Analysis} 
 (1) We find that samples from both sets suffering large domain shifts tend to have large differences in optimization directions, which will cause the optimization of $\omega$ in Eq.~\ref{meta} to be weakened by domain differences rather than the pseudo-label noise.
 As shown in Fig. \ref{dir}, despite using real labels in bi-level optimization, domain factors (\emph{e.g.} the spatial layout or image style)  will still bias the optimization direction.
 (2) We find that samples from both sets with inconsistent category distribution cannot produce intersecting gradients in Eq.~\ref{meta}, causing the corresponding gradient of $\omega$ to be 0 as well. Moreover, in actual optimization, it is unacceptable to use massive stable samples to perform multi-step gradient descent for Eq.~\ref{inner_1} due to huge computational and memory overhead. 
As a result, stable and unstable samples always suffer from mismatched class distributions, resulting in no gradient for missing classes.

In conclusion, it is reasonable to argue that matching highly correlated samples from two sets is critical for unbiased uncertain estimation.

\subsection{Stable Neighbor Denoising}
We propose to query the stable neighbor with similar domain properties for each unstable sample and compensate for missing categories for the neighbor, resulting in an unbiased estimate of $\omega$.
In the querying part, following the observation in Eq.~\ref{dir}, we specify domain factors, namely image style and spatial layout, as the proxy for querying.

Specifically, \textit{for image style}, a series of works \cite{Yang_2020_CVPR, li2022uncertainty} pointed out that the Fourier amplitude spectrum has a strong correlation with the image style. 
Thus, we adopt the flattened vector of the low-frequency Fourier amplitude spectrum as the style proxy, \emph{i.e.}, $\mathcal Q_{style} = Low (\mathcal A(\mathcal F(x)) $. \textit{For spatial layout}, \cite{10.1007/978-3-030-58568-6_26} shows that the row-wise and column-wise label statistical histogram vectors can represent the spatial distribution.
Thus, we use the pseudo-labels of both sets samples for computation as an alternative, \emph{i.e.}, $\mathcal Q_{layout} = [\mathcal Q_{layout}^{row}, \mathcal Q_{layout}^{col}]$, where $ Q_{layout}^{row, c} = \frac{{\sum_{row} \arg\max [\mathcal H(x_{t}^{l}  |  \Theta^{m})] = c} }{\sum_{row}\sum_{column} \arg\max [\mathcal H(x_{t}^{l}  |  \Theta^{m})] = c}  $, $c \in [0, 1,... C]$, and $C$ is the number of categories. With $\mathcal Q_{style} $ and  $\mathcal Q_{layout} $, we can retrieve the nearest stable neighbor $x^{i}_{neib} \in {\mathcal D}_{se} $ for the unstable sample $x^{i} \in {\mathcal D}_{ue} $ via the similarity between the proxy vectors.

The retrieved domain-associated neighbors may still suffer from missing categories, leading to problems described in the analysis (2).
Thus, we further propose to borrow the missing class objects from other stable samples and paste them to the neighbor as compensation.
However, the tail categories in the stable set ${\mathcal D}_{se}$ may be scarce, making the objects to be copied far less abundant.
To enrich the diversity of tail categories, we additionally maintain top-$k$ ranked stable category sets $ \ {\mathcal D}_{sec}^{c}=\{(x^{i}, \bar y^{i}) | x^{i} \in  \ {\mathcal D}_{t}, { \rm ES}^{i,m,c} \in  \top_k \}$ for each category $c$ as the supplement. 
We name the $ \ {\mathcal D}_{se}$ as a stable whole set and the $ \ {\mathcal D}_{sec}^{c} $  as a stable category set for distinction.
Subsequently, the nearest stable neighbor $x^{i}_{neib} \in {\mathcal D}_{se} $ can be compensated as follows,
\begin{equation} \label{comp}
\bar x^{i}_{neib}, \bar y^{i}_{neib} = {\rm CP}[(x^{i}_{neib}, \bar y^{i}_{neib}), (x_{mix}, \bar y_{mix}) \in {\mathcal D}_{sec}^{c}],
\end{equation}
where ${\rm CP}[\cdot, \cdot]$ is the copy-paste operation\cite{tranheden2021dacs} that copies objects from the latter to the former.

Through these efforts, we can use unbiased stable neighbors $\bar x^{i}_{neib}$ for Eq.~\ref{inner_1}  to denoise unstable samples in an unbiased way. Fig.~\ref{pipeline} shows the pipeline of the proposed SND and the corresponding is shown in Algorithm \ref{alg}.
Algorithm \ref{alg} implements our SND within the student-teacher framework \cite{sohn2020fixmatch} to align with existing methods. See detailed explanations and ablation in the Appendix.

\begin{algorithm}[t] 
\caption{The training step of SND.}
\label{alg}
\hspace*{0.02in} {\bf Input:}{ \  \  Target data $ {\mathcal D}_{t} $, source-trained segmentation network $ \mathcal H	$ with parameter $ \Theta $, max iteration $M$.} \\
\hspace*{0.02in} {\bf Output:} {Well-adapted segmentation model $ \mathcal H	$.}

\begin{algorithmic}[1]
\State Split $ {\mathcal D}_{t} $ into ${\mathcal D}_{se}$ and ${\mathcal D}_{ue}$ via vanilla self-training by Eq. \ref{es}, Eq. \ref{d_se} and Eq. \ref{d_ue} 
\State Initialize the teacher student model as $ \Theta_{tea}$ and $ \Theta_{stu}$
\For{$ m \gets 0 $ to $M-1$}
\State Sample batch samples $\{{x^{i}_t}\}^{B}_{i=1} \in {\mathcal D}_{ue}$  
\State \textbf{Inner loop Optimization}:
\State Retrieve nearest stable neighbor $(x^{i}_{neib}, y^{i}_{neib})\in {\mathcal D}_{se} $ for each unstable sample ${x^{i}_t}$ by $\mathcal Q_{style}$ and $\mathcal Q_{layout}$
\State Compensate $(x^{i}_{neib}, y^{i}_{neib})$ by Eq. \ref{comp}
\State Optimize $\Theta_{tea}$ by Eq.~\ref{meta} to obtain the uncertainty $\omega_{\star}$ and pseudo-label $\hat y_{t}$
\State \textbf{Outer loop Optimization}:
\State Update $ \Theta_{stu}$ by Eq. \ref{outer_1} with $\omega_{\star}$ and $\hat y_{t}$. 
\State Update $ \Theta_{tea}$ by exponential moving average

\EndFor\State{\bf end for}
\State {\bf Return} Adapted model $ \mathcal H	$ with $ \Theta_{tea}$.
\end{algorithmic}
\end{algorithm}

\begin{table*}[t]
  \centering
  \setlength{\abovecaptionskip}{0.1cm}
  \resizebox{\textwidth}{!}{
    \begin{tabular}{c|ccccccccccccccccccc|c}
    \toprule
          & \begin{sideways}road\end{sideways} & \begin{sideways}sidewalk\end{sideways} & \begin{sideways}Building\end{sideways} & \begin{sideways}Wall\end{sideways} & \begin{sideways}fence\end{sideways} & \begin{sideways}pole\end{sideways} & \begin{sideways}light\end{sideways} & \begin{sideways}sign\end{sideways} & \begin{sideways}vege.\end{sideways} & \begin{sideways}terrain\end{sideways} & \begin{sideways}sky\end{sideways} & \begin{sideways}person\end{sideways} & \begin{sideways}rider\end{sideways} & \begin{sideways}car\end{sideways} & \begin{sideways}truck\end{sideways} & \begin{sideways}bus\end{sideways} & \begin{sideways}train\end{sideways} & \begin{sideways}mbike\end{sideways} & \begin{sideways}bike\end{sideways} & mIoU \\
    \midrule
    \multicolumn{21}{c}{Source-free Synthetic-to-Real: GTA $\rightarrow$ Cityscapes (Val.)} \\
    \midrule
    HCL (NIPS'21)\cite{huang2021model}  & 92.6  & 54.6  & 82.8  & 33.2  & 26.2  & 39.8  & 38.1  & 31.9  & 84.5  & 38.6  & 85.3  & 61.3  & 30.2  & 85.4  & 33.1  & 41.6  & 14.4  & 27.3  & 44.0  & 49.7  \\
    SFDASEG (ICCV'21) $^{\dag}$  \cite{Kundu_2021_ICCV} & 91.7  & 53.4  & 86.1  & 37.6  & 32.1  & 37.4  & 38.2  & 35.6  & 86.7  & 48.5  & 89.9  & 62.6  & 34.3  & 87.2  & 51.0  & 50.8  & 4.2   & 42.7  & 53.9  & 53.4  \\
    DTST (CVPR'23) \cite{zhao2023towards} & 93.5  & 57.6  & 84.7  & 36.5  & 25.2  & 33.4  & \textbf{44.7} & 36.7  & 86.8  & 42.8  & 81.3  & 62.3  & 37.2  & 88.1  & 48.7  & \textbf{50.6} & 35.5  & 48.3  & 59.1  & 55.4  \\
    CrossMatch(ICCV'23)$^{\dag}$ \cite{yin2023crossmatch} & 95.1  & 67.8  & 87.7  & 51.3  & 41.5  & 36.3  & 47.4  & 51.3  & 87.8  & 47.8  & 87.3  & 67    & 34.2  & 87.5  & 41    & 51.8  & 0     & 42.6  & 46.4  & 56.4 \\
    CROTS(IJCV'24)\cite{luo2024crots} & 92    & 52.4  & 85.9  & 37.3  & 35.8  & 34.6  & 42.2  & 38.4  & 86.9  & 45.6  & 91.1  & 65.1  & 36.1  & 87.3  & 41.6  & 51.1  & 0     & 41.4  & 56.2  & 53.7 \\
    
    SND (Ours) & 93.0  & 54.0  & 84.6  & 35.6  & 30.3  & 31.0  & 41.9  & 41.6  & 87.6  & 44.6  & 86.4  & 62.6  & 38.5  & 87.5  & 48.7  & 42.9  & \textbf{36.6} & 49.5  & 58.7  & 55.6  \\
    DTST + SND (Ours) & \textbf{93.9} & \textbf{60.0} & \textbf{86.7} & \textbf{38.6} & \textbf{35.9} & \textbf{37.5} & 43.4  & \textbf{48.3} & \textbf{87.6} & \textbf{44.6} & \textbf{90.1} & \textbf{65.3} & \textbf{39.9} & \textbf{88.5} & \textbf{54.9} & 44.4  & 33.1  & \textbf{49.9} & \textbf{60.9} & \textbf{58.1} \\
    \midrule
    \midrule
    \multicolumn{21}{c}{Source-free Synthetic-to-Real: Synthia $\rightarrow$ Cityscapes (Val.)} \\
    \midrule
    HCL (NIPS'21) \cite{huang2021model} & 86.7  & 38.1  & 82.7  & 10    & 0.6   & 30.3  & 25.4  & 29.7  & 82.8  &  -    & 85.9  & 61.9  & 24.8  & 84.5  &    -   & 38.9  &  -     & 22.6  & 37.9  & 46.4  \\
    SFDASEG (ICCV'21) $^{\dag}$  \cite{Kundu_2021_ICCV} & \textbf{90.5} & \textbf{50.0} & 81.6  & 13.3  & 2.8   & 34.7  & 25.7  & 33.1  & 83.8  &     -  & \textbf{89.2} & \textbf{66.0} & \textbf{34.9} & 85.3  &      - & 53.4 &  -     & 46.1  & 46.6  & 52.3  \\
    DTST (CVPR'23) \cite{zhao2023towards} & 88.9  & 45.8  & 83.3  & 13.7  & 0.8   & 32.7  & 31.6  & 20.8  & 85.7  &    -   & 82.5  & 64.4  & 27.8  & \textbf{88.1} & -      & 50.9  &    -   & 37.6  & \textbf{57.3}  & 50.7  \\
    CrossMatch(ICCV'23)$^{\dag}$ \cite{yin2023crossmatch} & 91.5  & 55.5  & \textbf{85.4}  & \textbf{34.4}  & 8.3   & \textbf{40.8} & \textbf{40.0}    & 44.4  & 86.6  & - & 84.3  & 62.4  & 22.0    & \textbf{88.3}  & - & \textbf{60.0}    & - & 40.6  & 45.6   & 55.6  \\
    CROTS(IJCV'24)\cite{luo2024crots} & 89.4  & 41.6  & 82.7  & 15.1  & 1.2   & 34.7  & 33.7  & 25.7  & 83.7  & - & 87.9  & 66.6  & 34.6  & 85.4  & - & 45.9  & - & 43.5  & 49.6  & 51.3  \\
    SND (Ours) & 88.1  & 47.4  & 80.1  & 28.1  & \textbf{32.2} & 34.9  & 33.6  & 41.3  & 83.3  &    -   & 86.7  & 59.9  & 27.2  & 86.7  &      - & 48.1  &    -   & 36.2  & 52.5  & 54.1  \\
    DTST + SND (Ours) & 88.7  & 43.7  & 83.6  & 32.1  & 26.0  & 32.4  & 38.0 & \textbf{44.7}  & \textbf{87.2}  &  -   & 87.9  & 62.2  & \textbf{35.5} & 87.4  &  -   & 40.4  &  -   & \textbf{46.9} & 57.2  & \textbf{55.9}  \\ 
    \midrule
    \midrule
    \multicolumn{21}{c}{Source-free clean-Adverse-Weather: Cityscapes $\rightarrow$ ACDC (Test)} \\
    \midrule
    URMDA (CVPR2021) $^{\dag}$  \cite{fleuret2021uncertainty} & 74.1  & 25.6  & 43.5  & 19.5  & 24.3  & 25.9  & 48.6  & 43.2  & 66.2  & 27.5  & 79.8  & 45.0  & 20.9  & 70.7  & 36.8  & 35.9  & 32.7  & 27.7  & 29.9  & 40.9  \\
    HCL (NIPS2021) \cite{huang2021model} & 72.6  & 24.7  & 68.5  & 21.4  & 19.4  & 31.0  & 51.8  & 46.5  & 71.8  & 28.5  & 81.2  & 43.8  & 21.5  & 76.8  & 42.8  & 44.4  & 36.1  & 30.1  & 24.0  & 44.0  \\
    SFDASEG (ICCV21) $^{\dag}$  \cite{Kundu_2021_ICCV} & 73.5  & 29.0  & 70.7  & 19.7  & 21.9  & 36.1  & 53.2  & 51.4  & 72.0  & 31.1  & 85.3  & 41.5  & 26.2  & 76.4  & 41.9  & 46.1  & 39.8  & 33.3  & 32.0  & 46.4  \\
    DTST (CVPR2023) \cite{zhao2023towards} & 73.4  & 28.5  & 69.7  & 16.7  & 20.5  & 32.8  & 50.2  & 51.1  & 71.5  & 30.8  & 85.0  & 50.8  & 22.3  & 71.6  & 40.3  & 41.7  & 35.6  & 31.9  & 38.2  & 45.4  \\
    SND (Ours) & \textbf{74.8 } & 27.6  & 69.3  & \textbf{23.6} & \textbf{26.3} & 36.5  & 52.8  & 54.6  & 75.5  & 35.0  & 84.8  & 52.5  & 25.5  & 78.8  & 44.9  & 48.1  & 37.6  & 31.1  & 34.9  & 48.1  \\
    DTST + SND (Ours) & 73.4  & \textbf{29.4} & \textbf{70.9} & 22.0  & 25.1  & \textbf{38.5} & \textbf{54.6} & \textbf{55.5} & \textbf{77.7} & \textbf{35.2} & \textbf{86.6} & \textbf{53.4} & \textbf{27.4} & \textbf{80.9} & \textbf{45.6} & \textbf{49.0} & \textbf{41.0} & \textbf{36.7} & \textbf{40.4} & \textbf{49.6} \\
    \midrule
    \midrule
    \multicolumn{21}{c}{Source-free Open-compound: GTA $\rightarrow$ BDD100k (Test)} \\
    \midrule
    URMDA (CVPR'21) $^{\dag}$  \cite{fleuret2021uncertainty} & 83.9  & 38.3  & 78.7  & 9.6   & 7.3   & 29.1  & 11.1  & 4.9   & 70.7  & -      & 74.2  & 53.8  & 15.0  & 81.2  &   -    & 35.0  &   -    & 22.8  & 30.5  & 40.4  \\
    HCL (NIPS'21) \cite{huang2021model} & 88.6  & 39.2  & 81.0  & 8.2   & 7.9   & 28.4  & 11.4  & 5.7   & 71.0  & -      & 77.2  & 54.2  & 16.0  & 81.8  &  -     & 41.4  &    -   & 22.6  & 31.4  & 41.6  \\
    SFDASEG (ICCV'21) $^{\dag}$  \cite{Kundu_2021_ICCV} & \textbf{87.9} & 40.2  & \textbf{80.6 } & 13.1  & 8.2   & 30.2  & 22.8  & 17.1  & 71.1  &      - & 78.1  & 51.4  & 27.9  & 80.2  &  -     & \textbf{43.7} &    -   & 30.3  & 42.3  & 45.3  \\
    DTST (CVPR'23) \cite{zhao2023towards} & 83.1  & 39.9  & 64.9  & 8.9   & 14.5  & 29.5  & 27.0  & 27.1  & \textbf{71.9} &  -     & 83.2  & 52.9  & 31.3  & 74.7  &     -  & 41.1  &    -   & 30.0  & 42.1  & 45.2  \\
    SND (Ours) & 84.1  & 42.6  & 74.1  & 15.2  & 21.2  & 31.1  & 31.0  & 25.5  & 70.4  &    -   & 83.9  & 52.8  & 33.9  & 79.9  &    -   & 39.1  &     -  & 37.5  & 41.9  & 47.8  \\
    SND + DTST (Ours) & 86.5  & \textbf{44.4} & 77.3  & \textbf{21.3} & \textbf{22.9} & \textbf{32.4} & \textbf{33.0} & \textbf{27.4} & 69.6  &     -  & \textbf{86.7} & \textbf{54.3} & \textbf{34.3} & \textbf{82.1} &    -   & 39.4  &  -     & \textbf{38.3} & \textbf{42.4} & \textbf{49.5} \\
    \bottomrule    
    \end{tabular}%
    }
    \caption{Comparison of SND with state-of-the-art works on the tasks of source-free domain adaptation in semantic segmentation. The model is deeplab-v2 with ResNet101. The report metric is IoU($\%$). $^{\dag}$denotes using the specific network on the source side, \emph{e.g.}, SFDASEG using multiple heads, CrossMatch using two segmentation models with depth estimation. DTST+SND means using the minority class resampling strategy in DTST\cite{zhao2023towards}, as minority class adaptation is very challenging in source-free UDA.} 
  \label{tab:sfuda_r100}%
\vspace{-0.4cm}
\end{table*}

\section{Experiments}
\subsection{Datasets \& Setup}

\noindent \textbf{Datasets}.   GTA5 \cite{gta5_dataset} dataset provides 24,999 game-rendering urban scene images with a resolution of 1914×1024. 
SYNTHIA \cite{syhth_dataset} dataset includes 9,400 virtual images with a resolution of 1280×760. 
We use 19 and 16 common categories in these two datasets respectively as source data. Cityscapes \cite{Cordts_2016_CVPR} dataset contains 3,975 real urban scene images from 50 different cities in primarily Germany, with a resolution of 2048×1024.
ACDC \cite{SDV21} is the real-world dataset on adverse visual conditions, which comprises diverse weather scene images with a resolution of 1280×720.
BDD100K \cite{yu2020bdd100k} is a compound real-world dataset consists of 8000 training images and 1000 verification images with a resolution of 1280×720.


\noindent \textbf{Setup}. We conduct experiments using the Convolutional and Transformer structures respectively.
For Convolutional structures, we adopt the Deeplab-v2 \cite{deeplab_v2} with ResNet-101 \cite{He_2016_CVPR} as the segmentation model.
SGD optimizers are used for both inner and outer optimization, where the outer optimizer is set an initial learning rate ($\beta$) to $2.5 \times 10^{-4}$ with weight decay 0.01 while the inner optimizer is set a fixed learning rate ($\alpha$) of $0.01$.
The batch size is set as 4, and the framework is trained for 20,000 iterations on all SFUDA tasks.
For Transformer structures, we adopt the SegFormer \cite{xie2021segformer} with MiT-B5 \cite{xie2021segformer} .
AdamW \cite{loshchilov2017decoupled} and Adam \cite{kingma2014adam} optimizer are used for inner and outer optimization respectively, where the outer optimizer is set an initial learning rate ($\beta$) $6 \times 10^{-5}$ with weight decay 0.01, while the inner optimizer is set a fixed learning rate ($\alpha$) of $0.01$.
The batch size is set as 2 and the model is trained for 20,000 iterations on all tasks.
For set dividing, we use the stability (${\rm ES}$) evaluated at 10k ($\tau$) iterations and regard the top $5\%$ ($k\%$) ranked stable samples as the stable set for all adaptation tasks.

\subsection{Comparison with State-of-the-art Alternatives}
\noindent \textbf{Performance Comparison on SFUDA.} We compare our methods with the state-of-the-art approaches on source-free unsupervised domain adaptive semantic segmentation (SFUDA), including adapting to single and compound target domains. 
Table~\ref{tab:sfuda_r100} shows that the proposed SND achieves the best performance on all SFUDA tasks. Moreover, SND can be combined with other method (DTST~\cite{zhao2023towards}) and achieves further significant improvements. In the two synthetic-to-real SFUDA adaptation tasks, our approach (SND+DTST) surpasses the second-highest methods by $2.7\%$ and $3.1\%$, respectively. 
On real-to-real adaptation tasks, \textit{i.e.}, ACDC with various weather conditions, our approach (SND+DTST) still maintains excellent performance, improving the performance of the current best method (SFDASEG~\cite{Kundu_2021_ICCV}) by $3.2\%$.
When adapted to the more challenging open-compound domain BDD100k,  the proposed method  (SND+DTST) achieves a large performance gain, surpassing the second-highest method by $4.2\%$. The performance under various conditions proves that our method can effectively cope with the pseudo-label noise in complex environments, and alleviate the bias problem in SFUDA tasks.
%
%
%

\begin{table}[t]
  \centering
   \renewcommand{\arraystretch}{1.05}
  \resizebox{0.425\textwidth}{!}{
    \begin{tabular}{c|cccc|c}
    \toprule
    \multicolumn{6}{c}{Continual source-free adaptation: Cityscapes$\rightarrow$ACDC} \\
    \hline
      Time    & \multicolumn{5}{c}{$t$ $\longrightarrow$} \\
    \hline
          & Fog   & Night & Rain  & Snow & mIoU \\
    \hline
    Source model \cite{xie2021segformer} & 69.1  & 40.3  & 59.7  & 57.8  & 56.7  \\
    TENT $	^\ast$ \cite{wang2020tent} & 68.5  & 36.3  & 59.9  & 54.7  & 54.9  \\
    CoTTA  $	^\ast $ \cite{wang2022continual} & 70.4  & 41.6  & 63.9  & 60.8  & 59.2  \\
    HCL \cite{huang2021model}  & 70.0  & 39.9  & 63.7  & 61.2  & 58.7  \\
    SFDASEG $^{\dag}$ \cite{Kundu_2021_ICCV}  & 70.1  & 42.1  & 62.4  & 61.8  & 59.1 \\
    SND   & \textbf{72.1} & \textbf{43.1} & \textbf{66.3} & \textbf{65.6} & \textbf{61.8} \\
    \bottomrule
    \end{tabular}%
     }
 	\setlength{\abovecaptionskip}{0.05 cm}
  \caption{Comparison on the tasks of continual source-free domain adaptation semantic segmentation. The segmentation model is SegFormer with MiT-B5 as the backbone. $^\ast$means we use the target domain data for multiple round adaptations rather than one round in the original paper.}
  \label{csfda}%
  \vspace{-0.2cm}
\end{table}%

\begin{table*}[htbp]
  \centering
   \renewcommand{\arraystretch}{1.05}
  \resizebox{0.85\textwidth}{!}{    
    \begin{tabular}{c|cccccccc}
    \toprule
          & Unweighted & PE.\cite{wang2020tent}  & MCC.\cite{zheng2021rectifying}  & DAC. \cite{ Kundu_2021_ICCV}  & MPC.\cite{fleuret2021uncertainty}  & PD.\cite{zhang2021prototypical}  &  CIA.\cite{10005033}  & \textbf{SND}  \\
    \midrule
    GTA $\rightarrow$ Cityscapes   & 49.6  & 50.5 \textcolor[RGB]{18,220,168}{(+0.9)}  & 52.1 \textcolor[RGB]{18,220,168}{(+2.5)} & 51.7 \textcolor[RGB]{18,220,168}{(+2.1)}  & 51.6 \textcolor[RGB]{18,220,168}{(+2.0)} & 50.9 \textcolor[RGB]{18,220,168}{(+1.3)} & 49.8 \textcolor[RGB]{18,220,168}{(+0.2)}  & \textbf{55.6} \textbf{\textcolor[RGB]{18,220,168}{(+6.0)}}  \\
    Synthia $\rightarrow$ Cityscapes  & 48.2  & 49.5 \textcolor[RGB]{18,220,168}{(+1.3)}  & 52.9  \textcolor[RGB]{18,220,168}{(+2.7)}  & 49.4 \textcolor[RGB]{18,220,168}{(+1.2)}   & 51.5 \textcolor[RGB]{18,220,168}{(+3.3)}  & 49.7 \textcolor[RGB]{18,220,168}{(+1.5)} & 47.2 \textcolor{red}{(-1.0)}   & \textbf{54.1}  \textbf{\textcolor[RGB]{18,220,168}{(+5.9)}} \\
    Cityscapes $\rightarrow$ ACDC   & 43.1  & 41.3 \textcolor{red}{(-1.8)} & 45.1 \textcolor[RGB]{18,220,168}{(+2.0)}  & 42.9  \textcolor{red}{(-0.3)} & 43.0 \textcolor{red}{(-0.1)}  & 39.1 \textcolor{red}{(-4.0)}  & 39.2 \textcolor{red}{(-3.9)} & \textbf{48.1}   \textbf{\textcolor[RGB]{18,220,168}{(+5.0)}}  \\
    GTA $\rightarrow$ BDD100k   & 42.1  & 41.5 \textcolor{red}{(-0.6)}  & 44.8 \textcolor[RGB]{18,220,168}{(+2.7)}  & 41.2 \textcolor{red}{(-0.9)}  & 42.9 \textcolor[RGB]{18,220,168}{(+0.8)}  & 40.5 \textcolor{red}{(-1.6)} & 37.1 \textcolor{red}{(-5.0)}   & \textbf{47.8} \textbf{\textcolor[RGB]{18,220,168}{(+5.7)}} \\
    \bottomrule
    \end{tabular}%
     }
    \setlength{\abovecaptionskip}{0.05 cm}
   \caption{Ablation experiments on uncertainty estimation for source-free cross-domain segmentation tasks. We compared the following uncertainty estimation methods, probability entropy (PE.)\cite{wang2020tent}, multi-classifier consistency (MCC.)\cite{zheng2021rectifying}, data augmentation consistency (DAC.)\cite{ Kundu_2021_ICCV}, model perturbation consistency (MPC.)\cite{fleuret2021uncertainty}, prototype distance (PD.)\cite{zhang2021prototypical}, cross-image association\cite{10005033}(CIA.). 
`Unweighted' means original pseudo labels.}
  \label{abl_ue}%
  \vspace{-0.2cm}
\end{table*}%

\noindent \textbf{Performance Comparison on Continual SFUDA.} 
As real-world machine systems always operate in non-stationary environments, we also verify the effectiveness of our approach in the continual adaptation setting.
In Table \ref{csfda}, our method maintains good performance, with an average adaptability increase of $5.1\%$ in multiple domains under continual settings, showing obvious advantages compared with other methods.
It is verified that our method can better adapt to multi-domain environments and alleviate model degradation caused by domain bias and confirmation bias.

\vspace{-0.1cm}
\subsection{Ablation Study}
\noindent \textbf{Effectiveness of denoising module}. In Table \ref{abl_ue}, we ablate the denoising part, \emph{i.e.}, SND, and compare it with alternatives.
In two synthetic-to-real SFUDA experiments, SND achieves a performance improvement of nearly $6\%$, showing optimal competitiveness than other alternatives.
In multiple and compound adaptation tasks, SND produces a performance improvement of nearly $5\%$. On difficult transfer tasks, SND exhibits stable cross-domain adaptability, whereas other alternatives suffer from severe performance degradation.
The effectiveness of SND is mainly contributed to its accurate estimation of uncertain and comprehensive denoising capabilities.

\begin{table}[tbp]
  \centering
   \renewcommand{\arraystretch}{1.05}
  \resizebox{0.45\textwidth}{!}{
    \begin{tabular}{c|c|c}
    \toprule
    Division metric & GTA $\rightarrow$ Cityscapes   & Cityscapes $\rightarrow$ ACDC \\
    \midrule
    Image-level entropy \cite{Pan_2020_CVPR}  & 52.9  & 45.1 \\
    Patch-level entropy \cite{Tsai_2019_ICCV}  & 51.7  & 43.6 \\
    Image-level loss \cite{yang2022divide} & 52.7  & 44.1 \\
    Domain distance  \cite{Tzeng_2017_CVPR}  & 51.6  & 42.4 \\
    Stability  & \textbf{55.6} & \textbf{48.1} \\
    \bottomrule
    \end{tabular}%
     }
 	\setlength{\abovecaptionskip}{0.05 cm}
  \caption{Comparison of the subset division metrics.}
  \label{alb_dm}%
  \vspace{-0.2cm}
\end{table}%

\begin{table}[tbp]
  \centering
   \renewcommand{\arraystretch}{1.05}
  \resizebox{0.40\textwidth}{!}{
    \begin{tabular}{c|c|c}
    \toprule
    Query  & GTA $\rightarrow$ Cityscapes   & Cityscapes $\rightarrow$ ACDC \\
    \midrule
    Random & 54.2  & 44.7  \\
    $\mathcal Q_{layout}$  & 55.3  & 44.9  \\
    $\mathcal Q_{style}$  & 55.1  & 47.5  \\
    $\mathcal Q_{layout} + \mathcal Q_{style}$  & \textbf{55.6} & \textbf{48.1} \\
    \bottomrule
    \end{tabular}%
     }
 	\setlength{\abovecaptionskip}{0.05 cm}
  \caption{Ablation study of querying stable neighbor strategies. Random means randomly selecting stable samples from ${\mathcal D}_{se}$. }
  \label{alb_query}%
  \vspace{-0.4cm}
\end{table}%

\noindent \textbf{Effectiveness of the Stability Metric}.
In Table \ref{alb_dm}, we compare the proposed stability with other adaptability metrics for subset division, including regional entropy\cite{Pan_2020_CVPR, Tsai_2019_ICCV},  image-level loss \cite{yang2022divide}, and domain distance \cite{Tzeng_2017_CVPR}. 
Denoising using stability metrics shows obvious advantages on two adaptive tasks, with gains increased by $2.7\%$ and $3.0\%$ compared to the second competitor.
We think it contributes to the ability to screen more reliable samples to provide effective support for denoising.

\noindent \textbf{Effectiveness of Query Methods}.
In Table \ref{alb_query}, we verify the impact of different retrieval factors on adaptability.
It shows that gradually adding domain-related factors for retrieving stable neighbors can improve adaptability.
In special, on complex compound domain adaptation task, Cityscapes $\rightarrow$ ACDC, retrieving by domain factors shows greater improvements. This further verifies matching highly correlated neighbors is crucial for bi-level optimization.

\begin{figure*}[t]
    \begin{center}
    \centering 
    \includegraphics[width=0.95\textwidth, height=0.5\textwidth]{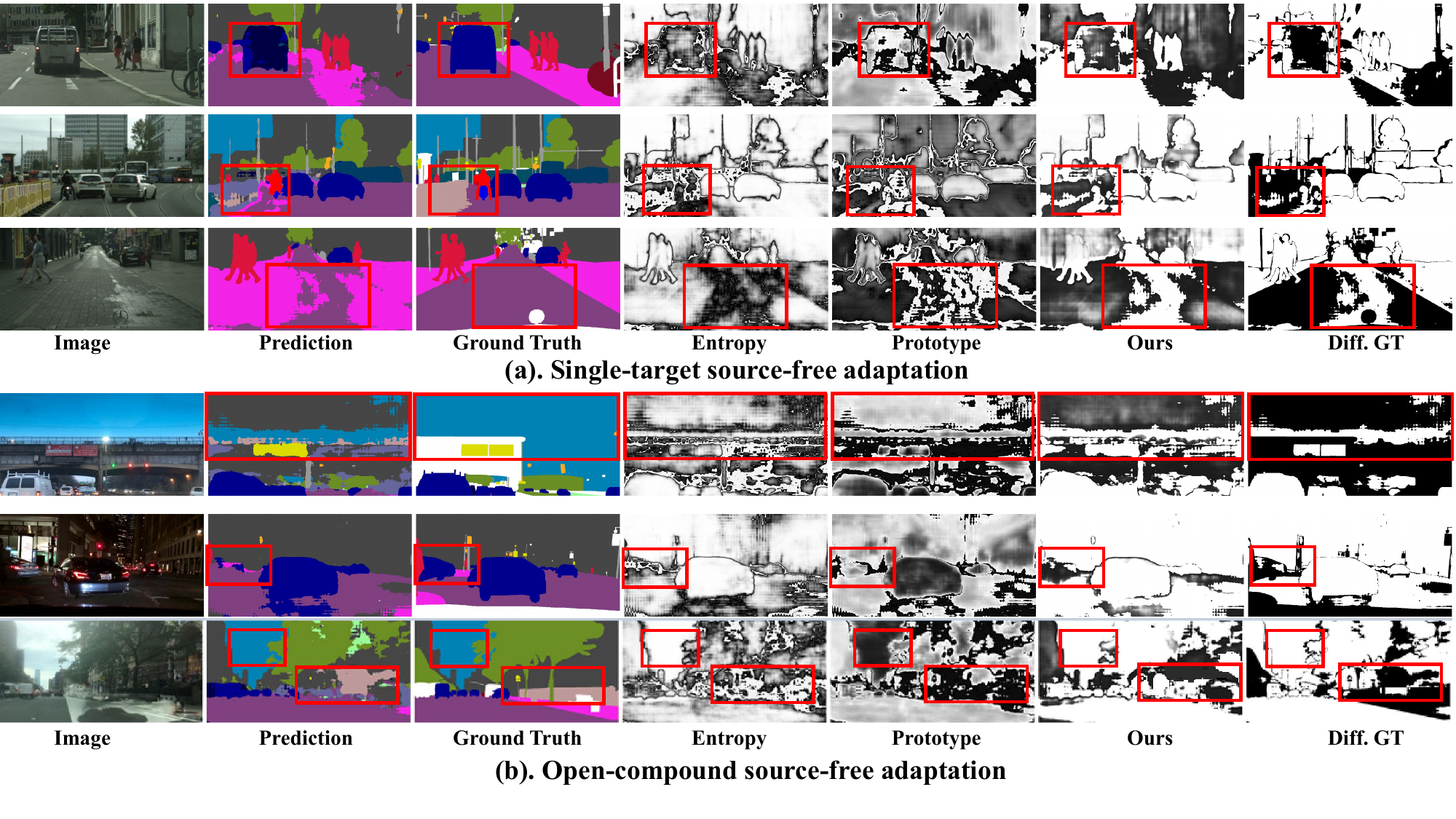}
    \end{center}
    \setlength{\abovecaptionskip}{-0.3 cm}
    \caption{Visualization of different uncertainty estimation results on both GTA $\rightarrow$ Cityscapes and GTA $\rightarrow$ BDD100k tasks. Diff.GT denotes the ground truth estimation mask. Entropy map is shown by probability entropy\cite{vu2019advent}.  Prototype map is shown by the difference between the Aspp classifier and prototype classifier \cite{zhang2021prototypical,shen2023diga}. } 
    \label{uncertain}
\vspace{-0.45cm}
\end{figure*}

\noindent \textbf{Effectiveness of Category Compensation}.
In Table \ref{alb_cc},  we verify the impact of category compensation on adaptability.
Using the stable set ${\mathcal D}_{se}$ as the copy object for compensation improves performance by $1.2\%$ and $0.8\%$ on two tasks respectively, verifying the impact of missing categories on denoising. 
Furthermore, adding category stable sets as replication objects can improve the performance by $1.4\%$ and $1.0\%$, showing that diversity compensation is more conducive to reducing the bias of bi-level optimization.
    
\begin{table}[tbp]
  \centering
   \renewcommand{\arraystretch}{1.05}
  \resizebox{0.45\textwidth}{!}{
    \begin{tabular}{c|c|c}
    \toprule
    Category compensation & GTA $\rightarrow$ Cityscapes   & Cityscapes $\rightarrow$ ACDC \\
    \midrule
    None  & 53.0  & 46.3  \\
    ${\mathcal D}_{se}$  & 54.2  & 47.1  \\
    ${\mathcal D}_{sec}$ & \textbf{55.6} & \textbf{48.1} \\
    \bottomrule
    \end{tabular}%
     }
 	\setlength{\abovecaptionskip}{0.05 cm}
  \caption{Variants of category compensation methods. `None' means not applying compensation. `${\mathcal D}_{se}$' and `${\mathcal D}_{sec}$' means we select samples from ${\mathcal D}_{se}$ and ${\mathcal D}_{sec}$ for compensation.}
  \label{alb_cc}%
  \vspace{-0.2cm}
\end{table}%

\subsection{Qualitative assessment}
\noindent \textbf{Visualization of the Learned Uncertainty Map }.
Fig. \ref{uncertain} presents the learned uncertain map $\omega$ on two SFUDA tasks. 
It can be seen that the uncertainty maps from probability entropy \cite{vu2019advent} and multi-classifier voting\cite{zhang2021prototypical,shen2023diga} contain a large amount of noise.
Moreover, in the harder open-compound task Cityscapes $\rightarrow$ ACDC,  the estimation bias of their methods is more obvious in difficult domain data, which will be a predisposing factor for error accumulation. In contrast, SND alleviates bias estimation and presents more reasonable uncertainty maps for different categories and domains. Moreover, our SND even can give accurate estimation against noise at the segmentation edges.

\begin{figure}[tbp]
    \begin{center}
    \centering \includegraphics[width=0.475\textwidth, height=0.22\textwidth]{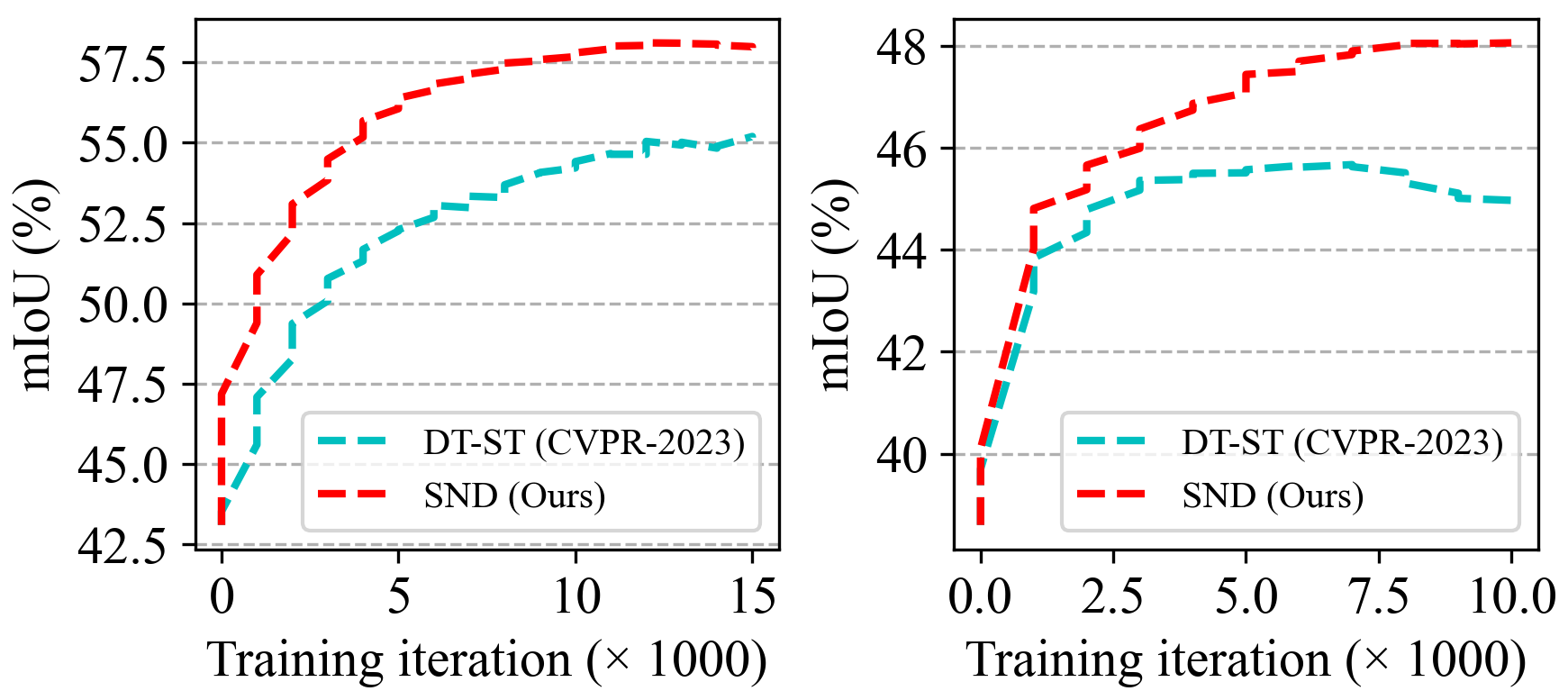}
    \end{center}
    \setlength{\abovecaptionskip}{-0.1 cm}
    \caption{Comparison of the mIoU score ($\%$) curve on the validation set during training on two transfer tasks.} 
    \setlength{\abovecaptionskip}{0.07 cm}
    \label{train_cur}
\vspace{-0.45cm}
\end{figure}

\noindent \textbf{Denoising Effect during Training}.
In Fig. \ref{train_cur}, we plot the mIoU scores on the validation during training to show the denoising effect. In single-target adaptation (a), compared with DTST\cite{zhao2023towards}, SND shows stronger denoising ability, allowing the model to achieve faster convergence speed during training and to obtain better performance.
In multi-target adaptation (b), DTST degrades in the later stages of training, which shows that more difficult adaptation tasks place higher requirements on denoising. 
In contrast, SND shows a better denoising effect and can effectively alleviate model degradation.

\subsection{Hyperparameter Sensitivity}
In Table~\ref{tau}, we analyze the sensitivity of the hyper-parameter top-$k \%$ and $\tau$ in Eq. \ref{es} on the two SFUDA tasks across three runs.
For $k$, we set the range from $1$ to $10$, because too small $k$ cannot select enough valuable samples, while too large $k$  will inject massive noise into the stable set.
On the two tasks, the fluctuation range of mIoU is within $0.5\%$ and $0.8\%$, showing that SND is not sensitive to $k$.
For $\tau$, we set the range from $2,000$ to $12,000$ to verify the impact of the stability evaluation iteration on SND for selecting stable samples. Results show that SND has very small fluctuations in performance, within $0.3\%$. This indicates that SND is not sensitive to $\tau$ and also verifies the observation in Fig.~\ref{stability} from the side.

\begin{table}[tbp]
  \centering
   \renewcommand{\arraystretch}{1.05}
  \resizebox{0.42\textwidth}{!}{
    \begin{tabular}{c|ccccc}
    \toprule
    $k $ (\%) & 1.0  & 2.5 & 5.0  & 7.5 & 10.0 \\
    \midrule
    GTA $\rightarrow$ Cityscapes   & 57.8 & 57.9  & \textbf{58.1}  & 57.8  & 57.8 \\
    Cityscapes $\rightarrow$ ACDC   & 49.9  & 50.2  & \textbf{50.7}  & 50.4  & 49.8 \\
    \midrule
    \midrule
    $\tau$ $(\times 1000)$ & 2     & 4     & 8     & 10    & 12 \\
    \midrule
    GTA $\rightarrow$ Cityscapes      & 57.9  & 58.0  & \textbf{58.1}  & 58.1  & 58.1  \\
     Cityscapes $\rightarrow$ ACDC  & 50.5  & 50.6  & \textbf{50.7}  & 50.7  & 50.7 \\
    \bottomrule
    \end{tabular}%
     }
 	\setlength{\abovecaptionskip}{0.08 cm}
  \caption{Sensitivity study of the hyper-parameter $k$ and $\tau$.}
  \label{tau}%
  \vspace{-0.6cm}
\end{table}%

\vspace{-0.1cm}
\section{Conclusion}
\vspace{-0.1cm}

In this work, we propose \emph{Stable Neighbor Denoising} to perform unbiased denoising for the SFUDA semantic segmentation tasks.
SND detects and suppresses noise in unstable samples by establishing the connection between the stable and unstable samples through bi-level optimization.
The proposed retrieval nearest neighbor strategy and category compensation strategy further reduce the bias of bi-level optimization, thereby achieving effective denoising.
Extensive experiments on different source-free adaptation scenarios, backbones, and ablations show that SND effectively estimates the noise of pseudo-labels and achieves state-of-the-art performance on all benchmarks.

{\small
\bibliographystyle{ieee_fullname}
\bibliography{ref}

\begin{thebibliography}{10}\itemsep=-1pt

\bibitem{araslanov2021self}
Nikita Araslanov and Stefan Roth.
\newblock Self-supervised augmentation consistency for adapting semantic segmentation.
\newblock In {\em Proceedings of the IEEE/CVF Conference on Computer Vision and Pattern Recognition}, pages 15384--15394, 2021.

\bibitem{Araslanov_2021_CVPR}
Nikita Araslanov and Stefan Roth.
\newblock Self-supervised augmentation consistency for adapting semantic segmentation.
\newblock In {\em Proceedings of the IEEE/CVF Conference on Computer Vision and Pattern Recognition (CVPR)}, pages 15384--15394, June 2021.

\bibitem{chen2022debiased}
Baixu Chen, Junguang Jiang, Ximei Wang, Pengfei Wan, Jianmin Wang, and Mingsheng Long.
\newblock Debiased self-training for semi-supervised learning.
\newblock {\em Advances in Neural Information Processing Systems}, 35:32424--32437, 2022.

\bibitem{deeplab_v2}
L. {Chen}, G. {Papandreou}, I. {Kokkinos}, K. {Murphy}, and A.~L. {Yuille}.
\newblock Deeplab: Semantic image segmentation with deep convolutional nets, atrous convolution, and fully connected crfs.
\newblock {\em IEEE Transactions on Pattern Analysis and Machine Intelligence}, 40(4):834--848, 2018.

\bibitem{chen2022deliberated}
Lin Chen, Zhixiang Wei, Xin Jin, Huaian Chen, Miao Zheng, Kai Chen, and Yi Jin.
\newblock Deliberated domain bridging for domain adaptive semantic segmentation.
\newblock {\em Advances in Neural Information Processing Systems}, 35:15105--15118, 2022.

\bibitem{MaxSquare_2019_ICCV}
Minghao Chen, Hongyang Xue, and Deng Cai.
\newblock Domain adaptation for semantic segmentation with maximum squares loss.
\newblock In {\em Proceedings of the IEEE/CVF International Conference on Computer Vision (ICCV)}, October 2019.

\bibitem{chen2019crdoco}
Yun-Chun Chen, Yen-Yu Lin, Ming-Hsuan Yang, and Jia-Bin Huang.
\newblock Crdoco: Pixel-level domain transfer with cross-domain consistency.
\newblock In {\em Proceedings of the IEEE/CVF conference on computer vision and pattern recognition}, pages 1791--1800, 2019.

\bibitem{chen2024liebn}
Ziheng Chen, Yue Song, Yunmei Liu, and Nicu Sebe.
\newblock A {Lie} group approach to {Riemannian} batch normalization.
\newblock In {\em The Twelfth International Conference on Learning Representations}, 2024.

\bibitem{cheng2020learning}
Hao Cheng, Zhaowei Zhu, Xingyu Li, Yifei Gong, Xing Sun, and Yang Liu.
\newblock Learning with instance-dependent label noise: A sample sieve approach.
\newblock {\em arXiv preprint arXiv:2010.02347}, 2020.

\bibitem{cheng2023adpl}
Yiting Cheng, Fangyun Wei, Jianmin Bao, Dong Chen, and Wenqiang Zhang.
\newblock Adpl: Adaptive dual path learning for domain adaptation of semantic segmentation.
\newblock {\em IEEE Transactions on Pattern Analysis and Machine Intelligence}, 2023.

\bibitem{choi2019self}
Jaehoon Choi, Taekyung Kim, and Changick Kim.
\newblock Self-ensembling with gan-based data augmentation for domain adaptation in semantic segmentation.
\newblock In {\em Proceedings of the IEEE/CVF International Conference on Computer Vision}, pages 6830--6840, 2019.

\bibitem{choi2021robustnet}
Sungha Choi, Sanghun Jung, Huiwon Yun, Joanne~T Kim, Seungryong Kim, and Jaegul Choo.
\newblock Robustnet: Improving domain generalization in urban-scene segmentation via instance selective whitening.
\newblock In {\em Proceedings of the IEEE/CVF Conference on Computer Vision and Pattern Recognition}, pages 11580--11590, 2021.

\bibitem{Cordts_2016_CVPR}
Marius Cordts, Mohamed Omran, Sebastian Ramos, Timo Rehfeld, Markus Enzweiler, Rodrigo Benenson, Uwe Franke, Stefan Roth, and Bernt Schiele.
\newblock The cityscapes dataset for semantic urban scene understanding.
\newblock In {\em Proceedings of the IEEE/CVF Conference on Computer Vision and Pattern Recognition (CVPR)}, June 2016.

\bibitem{ding2022source}
Ning Ding, Yixing Xu, Yehui Tang, Chao Xu, Yunhe Wang, and Dacheng Tao.
\newblock Source-free domain adaptation via distribution estimation.
\newblock In {\em Proceedings of the IEEE/CVF Conference on Computer Vision and Pattern Recognition}, pages 7212--7222, 2022.

\bibitem{fleuret2021uncertainty}
Francois Fleuret et~al.
\newblock Uncertainty reduction for model adaptation in semantic segmentation.
\newblock In {\em Proceedings of the IEEE/CVF Conference on Computer Vision and Pattern Recognition}, pages 9613--9623, 2021.

\bibitem{MetaCorrection}
Xiaoqing Guo, Chen Yang, Baopu Li, and Yixuan Yuan.
\newblock Metacorrection: Domain-aware meta loss correction for unsupervised domain adaptation in semantic segmentation.
\newblock In {\em Proceedings of the IEEE/CVF Conference on Computer Vision and Pattern Recognition (CVPR)}, pages 3927--3936, June 2021.

\bibitem{han2019deep}
Jiangfan Han, Ping Luo, and Xiaogang Wang.
\newblock Deep self-learning from noisy labels.
\newblock In {\em Proceedings of the IEEE/CVF international conference on computer vision}, pages 5138--5147, 2019.

\bibitem{he2021multi}
Jianzhong He, Xu Jia, Shuaijun Chen, and Jianzhuang Liu.
\newblock Multi-source domain adaptation with collaborative learning for semantic segmentation.
\newblock In {\em Proceedings of the IEEE/CVF Conference on Computer Vision and Pattern Recognition}, pages 11008--11017, 2021.

\bibitem{He_2016_CVPR}
Kaiming He, Xiangyu Zhang, Shaoqing Ren, and Jian Sun.
\newblock Deep residual learning for image recognition.
\newblock In {\em Proceedings of the IEEE/CVF Conference on Computer Vision and Pattern Recognition (CVPR)}, June 2016.

\bibitem{hoffman2018cycada}
Judy Hoffman, Eric Tzeng, Taesung Park, Jun-Yan Zhu, Phillip Isola, Kate Saenko, Alexei Efros, and Trevor Darrell.
\newblock Cycada: Cycle-consistent adversarial domain adaptation.
\newblock In {\em International conference on machine learning}, pages 1989--1998. Pmlr, 2018.

\bibitem{hoyer2022daformer}
Lukas Hoyer, Dengxin Dai, and Luc Van~Gool.
\newblock Daformer: Improving network architectures and training strategies for domain-adaptive semantic segmentation.
\newblock In {\em Proceedings of the IEEE/CVF Conference on Computer Vision and Pattern Recognition}, pages 9924--9935, 2022.

\bibitem{hoyer2022hrda}
Lukas Hoyer, Dengxin Dai, and Luc Van~Gool.
\newblock Hrda: Context-aware high-resolution domain-adaptive semantic segmentation.
\newblock In {\em European Conference on Computer Vision}, pages 372--391. Springer, 2022.

\bibitem{hoyer2023mic}
Lukas Hoyer, Dengxin Dai, Haoran Wang, and Luc Van~Gool.
\newblock Mic: Masked image consistency for context-enhanced domain adaptation.
\newblock In {\em Proceedings of the IEEE/CVF Conference on Computer Vision and Pattern Recognition}, pages 11721--11732, 2023.

\bibitem{huang2021fsdr}
Jiaxing Huang, Dayan Guan, Aoran Xiao, and Shijian Lu.
\newblock Fsdr: Frequency space domain randomization for domain generalization.
\newblock In {\em Proceedings of the IEEE/CVF Conference on Computer Vision and Pattern Recognition}, pages 6891--6902, 2021.

\bibitem{huang2021model}
Jiaxing Huang, Dayan Guan, Aoran Xiao, and Shijian Lu.
\newblock Model adaptation: Historical contrastive learning for unsupervised domain adaptation without source data.
\newblock {\em Advances in Neural Information Processing Systems}, 34:3635--3649, 2021.

\bibitem{izmailov2018averaging}
Pavel Izmailov, Dmitrii Podoprikhin, Timur Garipov, Dmitry Vetrov, and Andrew~Gordon Wilson.
\newblock Averaging weights leads to wider optima and better generalization.
\newblock {\em arXiv preprint arXiv:1803.05407}, 2018.

\bibitem{jing2022variational}
Mengmeng Jing, Xiantong Zhen, Jingjing Li, and Cees Snoek.
\newblock Variational model perturbation for source-free domain adaptation.
\newblock {\em Advances in Neural Information Processing Systems}, 35:17173--17187, 2022.

\bibitem{kingma2014adam}
Diederik~P Kingma and Jimmy Ba.
\newblock Adam: A method for stochastic optimization.
\newblock {\em arXiv preprint arXiv:1412.6980}, 2014.

\bibitem{Kundu_2021_ICCV}
Jogendra~Nath Kundu, Akshay Kulkarni, Amit Singh, Varun Jampani, and R.~Venkatesh Babu.
\newblock Generalize then adapt: Source-free domain adaptive semantic segmentation.
\newblock In {\em Proceedings of the IEEE/CVF International Conference on Computer Vision (ICCV)}, pages 7046--7056, October 2021.

\bibitem{kundu2021generalize}
Jogendra~Nath Kundu, Akshay Kulkarni, Amit Singh, Varun Jampani, and R~Venkatesh Babu.
\newblock Generalize then adapt: Source-free domain adaptive semantic segmentation.
\newblock In {\em Proceedings of the IEEE/CVF International Conference on Computer Vision}, pages 7046--7056, 2021.

\bibitem{lee2022fifo}
Sohyun Lee, Taeyoung Son, and Suha Kwak.
\newblock Fifo: Learning fog-invariant features for foggy scene segmentation.
\newblock In {\em Proceedings of the IEEE/CVF Conference on Computer Vision and Pattern Recognition}, pages 18911--18921, 2022.

\bibitem{10.1007/978-3-030-58568-6_26}
Guangrui Li, Guoliang Kang, Wu Liu, Yunchao Wei, and Yi Yang.
\newblock Content-consistent matching for domain adaptive semantic segmentation.
\newblock In Andrea Vedaldi, Horst Bischof, Thomas Brox, and Jan-Michael Frahm, editors, {\em Proceedings of the European Conference on Computer Vision (ECCV)}, pages 440--456, 2020.

\bibitem{li2022expansion}
Jinlong Li, Zequn Jie, Xu Wang, Xiaolin Wei, and Lin Ma.
\newblock Expansion and shrinkage of localization for weakly-supervised semantic segmentation.
\newblock {\em Advances in Neural Information Processing Systems}, 35:16037--16051, 2022.

\bibitem{li2022weakly}
Jinlong Li, Zequn Jie, Xu Wang, Yu Zhou, Xiaolin Wei, and Lin Ma.
\newblock Weakly supervised semantic segmentation via progressive patch learning.
\newblock {\em IEEE Transactions on multimedia}, 2022.

\bibitem{li2022class}
Ruihuang Li, Shuai Li, Chenhang He, Yabin Zhang, Xu Jia, and Lei Zhang.
\newblock Class-balanced pixel-level self-labeling for domain adaptive semantic segmentation.
\newblock In {\em Proceedings of the IEEE/CVF Conference on Computer Vision and Pattern Recognition}, pages 11593--11603, 2022.

\bibitem{li2022uncertainty}
Xiaotong Li, Yongxing Dai, Yixiao Ge, Jun Liu, Ying Shan, and LINGYU DUAN.
\newblock Uncertainty modeling for out-of-distribution generalization.
\newblock In {\em International Conference on Learning Representations}, 2022.

\bibitem{liang2021source}
Jian Liang, Dapeng Hu, Yunbo Wang, Ran He, and Jiashi Feng.
\newblock Source data-absent unsupervised domain adaptation through hypothesis transfer and labeling transfer.
\newblock {\em IEEE Transactions on Pattern Analysis and Machine Intelligence}, 44(11):8602--8617, 2021.

\bibitem{liu2021source}
Yuang Liu, Wei Zhang, and Jun Wang.
\newblock Source-free domain adaptation for semantic segmentation.
\newblock In {\em Proceedings of the IEEE/CVF Conference on Computer Vision and Pattern Recognition}, pages 1215--1224, 2021.

\bibitem{loshchilov2017decoupled}
Ilya Loshchilov and Frank Hutter.
\newblock Decoupled weight decay regularization.
\newblock {\em arXiv preprint arXiv:1711.05101}, 2017.

\bibitem{luo2024crots}
Xin Luo, Wei Chen, Zhengfa Liang, Longqi Yang, Siwei Wang, and Chen Li.
\newblock Crots: Cross-domain teacher--student learning for source-free domain adaptive semantic segmentation.
\newblock {\em International Journal of Computer Vision}, 132(1):20--39, 2024.

\bibitem{Luo2019TakingAC}
Yawei Luo, L. Zheng, T. Guan, Junqing Yu, and Y. Yang.
\newblock Taking a closer look at domain shift: Category-level adversaries for semantics consistent domain adaptation.
\newblock {\em Proceedings of the IEEE/CVF Conference on Computer Vision and Pattern Recognition (CVPR)}, pages 2502--2511, 2019.

\bibitem{ma2020normalized}
Xingjun Ma, Hanxun Huang, Yisen Wang, Simone Romano, Sarah Erfani, and James Bailey.
\newblock Normalized loss functions for deep learning with noisy labels.
\newblock In {\em International conference on machine learning}, pages 6543--6553. PMLR, 2020.

\bibitem{ma2023delving}
Yanbiao Ma, Licheng Jiao, Fang Liu, Yuxin Li, Shuyuan Yang, and Xu Liu.
\newblock Delving into semantic scale imbalance.
\newblock In {\em The Eleventh International Conference on Learning Representations}, 2023.

\bibitem{Ma_2023_CVPR}
Yanbiao Ma, Licheng Jiao, Fang Liu, Shuyuan Yang, Xu Liu, and Lingling Li.
\newblock Curvature-balanced feature manifold learning for long-tailed classification.
\newblock In {\em Proceedings of the IEEE/CVF Conference on Computer Vision and Pattern Recognition (CVPR)}, pages 15824--15835, June 2023.

\bibitem{Pan_2020_CVPR}
Fei Pan, Inkyu Shin, Francois Rameau, Seokju Lee, and In~So Kweon.
\newblock Unsupervised intra-domain adaptation for semantic segmentation through self-supervision.
\newblock In {\em Proceedings of the IEEE/CVF Conference on Computer Vision and Pattern Recognition (CVPR)}, June 2020.

\bibitem{pan2023cloud}
Mingjie Pan, Rongyu Zhang, Zijian Ling, Yulu Gan, Lingran Zhao, Jiaming Liu, and Shanghang Zhang.
\newblock Cloud-device collaborative adaptation to continual changing environments in the real-world.
\newblock In {\em Proceedings of the IEEE/CVF Conference on Computer Vision and Pattern Recognition}, pages 12157--12166, 2023.

\bibitem{pan2009survey}
Sinno~Jialin Pan and Qiang Yang.
\newblock A survey on transfer learning.
\newblock {\em IEEE Transactions on knowledge and data engineering}, 22(10):1345--1359, 2009.

\bibitem{Pham_2021_CVPR}
Hieu Pham, Zihang Dai, Qizhe Xie, and Quoc~V. Le.
\newblock Meta pseudo labels.
\newblock In {\em Proceedings of the IEEE/CVF Conference on Computer Vision and Pattern Recognition (CVPR)}, pages 11557--11568, June 2021.

\bibitem{pu2022meta}
Nan Pu, Yu Liu, Wei Chen, Erwin~M Bakker, and Michael~S Lew.
\newblock Meta reconciliation normalization for lifelong person re-identification.
\newblock In {\em Proceedings of the 30th ACM International Conference on Multimedia}, pages 541--549, 2022.

\bibitem{10187687}
Nan Pu, Zhun Zhong, Nicu Sebe, and Michael~S. Lew.
\newblock A memorizing and generalizing framework for lifelong person re-identification.
\newblock {\em IEEE Transactions on Pattern Analysis and Machine Intelligence}, 45(11):13567--13585, 2023.

\bibitem{gta5_dataset}
Stephan~R. Richter, Vibhav Vineet, Stefan Roth, and Vladlen Koltun.
\newblock Playing for data: Ground truth from computer games.
\newblock In Bastian Leibe, Jiri Matas, Nicu Sebe, and Max Welling, editors, {\em Proceedings of the European Conference on Computer Vision (ECCV)}, pages 102--118. Springer International Publishing, 2016.

\bibitem{syhth_dataset}
German Ros, Laura Sellart, Joanna Materzynska, David Vazquez, and Antonio~M. Lopez.
\newblock The synthia dataset: A large collection of synthetic images for semantic segmentation of urban scenes.
\newblock In {\em Proceedings of the IEEE/CVF Conference on Computer Vision and Pattern Recognition (CVPR)}, June 2016.

\bibitem{sakaridis2019guided}
Christos Sakaridis, Dengxin Dai, and Luc~Van Gool.
\newblock Guided curriculum model adaptation and uncertainty-aware evaluation for semantic nighttime image segmentation.
\newblock In {\em Proceedings of the IEEE/CVF International Conference on Computer Vision}, pages 7374--7383, 2019.

\bibitem{sakaridis2020map}
Christos Sakaridis, Dengxin Dai, and Luc Van~Gool.
\newblock Map-guided curriculum domain adaptation and uncertainty-aware evaluation for semantic nighttime image segmentation.
\newblock {\em IEEE Transactions on Pattern Analysis and Machine Intelligence}, 44(6):3139--3153, 2020.

\bibitem{SDV21}
Christos Sakaridis, Dengxin Dai, and Luc Van~Gool.
\newblock {ACDC}: The adverse conditions dataset with correspondences for semantic driving scene understanding.
\newblock In {\em Proceedings of the IEEE/CVF International Conference on Computer Vision (ICCV)}, October 2021.

\bibitem{shen2023diga}
Fengyi Shen, Akhil Gurram, Ziyuan Liu, He Wang, and Alois Knoll.
\newblock Diga: Distil to generalize and then adapt for domain adaptive semantic segmentation.
\newblock In {\em Proceedings of the IEEE/CVF Conference on Computer Vision and Pattern Recognition}, pages 15866--15877, 2023.

\bibitem{sohn2020fixmatch}
Kihyuk Sohn, David Berthelot, Nicholas Carlini, Zizhao Zhang, Han Zhang, Colin~A Raffel, Ekin~Dogus Cubuk, Alexey Kurakin, and Chun-Liang Li.
\newblock Fixmatch: Simplifying semi-supervised learning with consistency and confidence.
\newblock {\em Advances in neural information processing systems}, 33:596--608, 2020.

\bibitem{tranheden2021dacs}
Wilhelm Tranheden, Viktor Olsson, Juliano Pinto, and Lennart Svensson.
\newblock Dacs: Domain adaptation via cross-domain mixed sampling.
\newblock In {\em Proceedings of the IEEE/CVF Winter Conference on Applications of Computer Vision}, pages 1379--1389, 2021.

\bibitem{Tsai_2019_ICCV}
Yi-Hsuan Tsai, Kihyuk Sohn, Samuel Schulter, and Manmohan Chandraker.
\newblock Domain adaptation for structured output via discriminative patch representations.
\newblock In {\em Proceedings of the IEEE/CVF International Conference on Computer Vision (ICCV)}, October 2019.

\bibitem{Tzeng_2017_CVPR}
Eric Tzeng, Judy Hoffman, Kate Saenko, and Trevor Darrell.
\newblock Adversarial discriminative domain adaptation.
\newblock In {\em Proceedings of the IEEE/CVFs Conference on Computer Vision and Pattern Recognition (CVPR)}, July 2017.

\bibitem{vu2019advent}
Tuan-Hung Vu, Himalaya Jain, Maxime Bucher, Matthieu Cord, and Patrick P{\'e}rez.
\newblock Advent: Adversarial entropy minimization for domain adaptation in semantic segmentation.
\newblock In {\em Proceedings of the IEEE/CVF conference on computer vision and pattern recognition}, pages 2517--2526, 2019.

\bibitem{wang2020tent}
Dequan Wang, Evan Shelhamer, Shaoteng Liu, Bruno Olshausen, and Trevor Darrell.
\newblock Tent: Fully test-time adaptation by entropy minimization.
\newblock {\em arXiv preprint arXiv:2006.10726}, 2020.

\bibitem{FADA_ECCV_2020}
Haoran Wang, Tong Shen, Wei Zhang, Ling-Yu Duan, and Tao Mei.
\newblock Classes matter: A fine-grained adversarial approach to cross-domain semantic segmentation.
\newblock In Andrea Vedaldi, Horst Bischof, Thomas Brox, and Jan-Michael Frahm, editors, {\em Proceedings of the European Conference on Computer Vision (ECCV)}, pages 642--659, 2020.

\bibitem{wang2022continual}
Qin Wang, Olga Fink, Luc Van~Gool, and Dengxin Dai.
\newblock Continual test-time domain adaptation.
\newblock In {\em Proceedings of the IEEE/CVF Conference on Computer Vision and Pattern Recognition}, pages 7201--7211, 2022.

\bibitem{10181233}
Shuang Wang, Qi Zang, Dong Zhao, Chaowei Fang, Dou Quan, Yutong Wan, Yanhe Guo, and Licheng Jiao.
\newblock Select, purify, and exchange: A multisource unsupervised domain adaptation method for building extraction.
\newblock {\em IEEE Transactions on Neural Networks and Learning Systems}, pages 1--15, 2023.

\bibitem{9961139}
Shuang Wang, Dong Zhao, Chi Zhang, Yuwei Guo, Qi Zang, Yu Gu, Yi Li, and Licheng Jiao.
\newblock Cluster alignment with target knowledge mining for unsupervised domain adaptation semantic segmentation.
\newblock {\em IEEE Transactions on Image Processing}, 31:7403--7418, 2022.

\bibitem{wang2019symmetric}
Yisen Wang, Xingjun Ma, Zaiyi Chen, Yuan Luo, Jinfeng Yi, and James Bailey.
\newblock Symmetric cross entropy for robust learning with noisy labels.
\newblock In {\em Proceedings of the IEEE/CVF international conference on computer vision}, pages 322--330, 2019.

\bibitem{10005033}
Linshan Wu, Leyuan Fang, Xingxin He, Min He, Jiayi Ma, and Zhun Zhong.
\newblock Querying labeled for unlabeled: Cross-image semantic consistency guided semi-supervised semantic segmentation.
\newblock {\em IEEE Transactions on Pattern Analysis and Machine Intelligence}, 45(7):8827--8844, 2023.

\bibitem{wu2021dannet}
Xinyi Wu, Zhenyao Wu, Hao Guo, Lili Ju, and Song Wang.
\newblock Dannet: A one-stage domain adaptation network for unsupervised nighttime semantic segmentation.
\newblock In {\em Proceedings of the IEEE/CVF Conference on Computer Vision and Pattern Recognition}, pages 15769--15778, 2021.

\bibitem{wu2021learning}
Yichen Wu, Jun Shu, Qi Xie, Qian Zhao, and Deyu Meng.
\newblock Learning to purify noisy labels via meta soft label corrector.
\newblock In {\em Proceedings of the AAAI Conference on Artificial Intelligence}, pages 10388--10396, 2021.

\bibitem{xie2023sepico}
Binhui Xie, Shuang Li, Mingjia Li, Chi~Harold Liu, Gao Huang, and Guoren Wang.
\newblock Sepico: Semantic-guided pixel contrast for domain adaptive semantic segmentation.
\newblock {\em IEEE Transactions on Pattern Analysis and Machine Intelligence}, 2023.

\bibitem{xie2021segformer}
Enze Xie, Wenhai Wang, Zhiding Yu, Anima Anandkumar, Jose~M Alvarez, and Ping Luo.
\newblock Segformer: Simple and efficient design for semantic segmentation with transformers.
\newblock {\em Advances in Neural Information Processing Systems}, 34:12077--12090, 2021.

\bibitem{ijcv_org}
Feng Xue, Yicong Chang, Tianxi Wang, Yu Zhou, and Anlong Ming.
\newblock Indoor obstacle discovery on reflective ground via monocular camera.
\newblock {\em International Journal of Computer Vision (IJCV)}, pages 1573--1405, 2023.

\bibitem{yang2022divide}
Jianfei Yang, Xiangyu Peng, Kai Wang, Zheng Zhu, Jiashi Feng, Lihua Xie, and Yang You.
\newblock Divide to adapt: Mitigating confirmation bias for domain adaptation of black-box predictors.
\newblock {\em arXiv preprint arXiv:2205.14467}, 2022.

\bibitem{yang2021generalized}
Shiqi Yang, Yaxing Wang, Joost van~de Weijer, Luis Herranz, and Shangling Jui.
\newblock Generalized source-free domain adaptation.
\newblock In {\em Proceedings of the IEEE/CVF International Conference on Computer Vision}, pages 8978--8987, 2021.

\bibitem{yang2023exploring}
Senqiao Yang, Jiarui Wu, Jiaming Liu, Xiaoqi Li, Qizhe Zhang, Mingjie Pan, and Shanghang Zhang.
\newblock Exploring sparse visual prompt for cross-domain semantic segmentation.
\newblock {\em arXiv preprint arXiv:2303.09792}, 2023.

\bibitem{Yang_2020_CVPR}
Yanchao Yang and Stefano Soatto.
\newblock Fda: Fourier domain adaptation for semantic segmentation.
\newblock In {\em Proceedings of the IEEE/CVF Conference on Computer Vision and Pattern Recognition (CVPR)}, June 2020.

\bibitem{yi2023source}
Li Yi, Gezheng Xu, Pengcheng Xu, Jiaqi Li, Ruizhi Pu, Charles Ling, A~Ian McLeod, and Boyu Wang.
\newblock When source-free domain adaptation meets learning with noisy labels.
\newblock {\em arXiv preprint arXiv:2301.13381}, 2023.

\bibitem{yin2023crossmatch}
Yifang Yin, Wenmiao Hu, Zhenguang Liu, Guanfeng Wang, Shili Xiang, and Roger Zimmermann.
\newblock Crossmatch: Source-free domain adaptive semantic segmentation via cross-modal consistency training.
\newblock In {\em Proceedings of the IEEE/CVF International Conference on Computer Vision}, pages 21786--21796, 2023.

\bibitem{yu2020bdd100k}
Fisher Yu, Haofeng Chen, Xin Wang, Wenqi Xian, Yingying Chen, Fangchen Liu, Vashisht Madhavan, and Trevor Darrell.
\newblock Bdd100k: A diverse driving dataset for heterogeneous multitask learning.
\newblock In {\em Proceedings of the IEEE/CVF conference on computer vision and pattern recognition}, pages 2636--2645, 2020.

\bibitem{yue2019domain}
Xiangyu Yue, Yang Zhang, Sicheng Zhao, Alberto Sangiovanni-Vincentelli, Kurt Keutzer, and Boqing Gong.
\newblock Domain randomization and pyramid consistency: Simulation-to-real generalization without accessing target domain data.
\newblock In {\em Proceedings of the IEEE/CVF International Conference on Computer Vision}, pages 2100--2110, 2019.

\bibitem{zhang2021prototypical}
Pan Zhang, Bo Zhang, Ting Zhang, Dong Chen, Yong Wang, and Fang Wen.
\newblock Prototypical pseudo label denoising and target structure learning for domain adaptive semantic segmentation.
\newblock In {\em Proceedings of the IEEE/CVF Conference on Computer Vision and Pattern Recognition}, pages 12414--12424, 2021.

\bibitem{zhang2022divide}
Ziyi Zhang, Weikai Chen, Hui Cheng, Zhen Li, Siyuan Li, Liang Lin, and Guanbin Li.
\newblock Divide and contrast: Source-free domain adaptation via adaptive contrastive learning.
\newblock {\em Advances in Neural Information Processing Systems}, 35:5137--5149, 2022.

\bibitem{zhang2018generalized}
Zhilu Zhang and Mert Sabuncu.
\newblock Generalized cross entropy loss for training deep neural networks with noisy labels.
\newblock {\em Advances in neural information processing systems}, 31, 2018.

\bibitem{zhao2023towards}
Dong Zhao, Shuang Wang, Qi Zang, Dou Quan, Xiutiao Ye, and Licheng Jiao.
\newblock Towards better stability and adaptability: Improve online self-training for model adaptation in semantic segmentation.
\newblock In {\em Proceedings of the IEEE/CVF Conference on Computer Vision and Pattern Recognition}, pages 11733--11743, 2023.

\bibitem{Zhao_2023_ICCV}
Dong Zhao, Shuang Wang, Qi Zang, Dou Quan, Xiutiao Ye, Rui Yang, and Licheng Jiao.
\newblock Learning pseudo-relations for cross-domain semantic segmentation.
\newblock In {\em Proceedings of the IEEE/CVF International Conference on Computer Vision (ICCV)}, pages 19191--19203, October 2023.

\bibitem{zhao2023semantic}
Dong Zhao, Ruizhi Yang, Shuang Wang, Qi Zang, Yang Hu, Licheng Jiao, Nicu Sebe, and Zhun Zhong.
\newblock Semantic connectivity-driven pseudo-labeling for cross-domain segmentation, 2023.

\bibitem{zhao2022style}
Yuyang Zhao, Zhun Zhong, Na Zhao, Nicu Sebe, and Gim~Hee Lee.
\newblock Style-hallucinated dual consistency learning for domain generalized semantic segmentation.
\newblock In {\em European Conference on Computer Vision}, pages 535--552. Springer, 2022.

\bibitem{zheng2021rectifying}
Zhedong Zheng and Yi Yang.
\newblock Rectifying pseudo label learning via uncertainty estimation for domain adaptive semantic segmentation.
\newblock {\em International Journal of Computer Vision}, 129(4):1106--1120, 2021.

\bibitem{zou2018unsupervised}
Yang Zou, Zhiding Yu, BVK Kumar, and Jinsong Wang.
\newblock Unsupervised domain adaptation for semantic segmentation via class-balanced self-training.
\newblock In {\em Proceedings of the European conference on computer vision (ECCV)}, pages 289--305, 2018.

\end{thebibliography}
}

\end{document}